\documentclass[sigconf]{acmart}

\AtBeginDocument{%
  \providecommand\BibTeX{{%
    \normalfont B\kern-0.5em{\scshape i\kern-0.25em b}\kern-0.8em\TeX}}}

\setcopyright{acmcopyright}
\copyrightyear{2023}
\acmYear{2023}
\setcopyright{acmlicensed}\acmConference[MM '23]{Proceedings of the 31st
ACM International Conference on Multimedia}{October 29-November 3,
2023}{Ottawa, ON, Canada}
\acmBooktitle{Proceedings of the 31st ACM International Conference on
Multimedia (MM '23), October 29-November 3, 2023, Ottawa, ON, Canada}
\acmPrice{15.00}
\acmDOI{10.1145/3581783.3612096}
\acmISBN{979-8-4007-0108-5/23/10}

\usepackage{xcolor}
\usepackage{latexsym}
\usepackage[caption=false,font=normalsize,labelfont=sf,textfont=sf]{subfig}

\usepackage[utf8]{inputenc}

\usepackage{microtype}
\usepackage{graphicx}
\usepackage{tikz}
\usetikzlibrary{patterns,shapes,snakes,calendar,calc,backgrounds,folding,arrows,positioning,math}
\usepackage{calc}
\usepackage{comment}
\usepackage{amsmath,bm} % define this before the line numbering.
\usepackage{pgfplots}
\usepackage{multirow}
\usepackage{paralist}
\usepackage{booktabs}
\usepackage[noend]{algpseudocode}
\usepackage{bbding}
\usepackage{colortbl}
\usepackage{tabularx}
\newcommand{\paratitle}[1]{\vspace{1.5ex}\noindent \textbf{#1}}

\begin{document}

\title[ Holistic Spatio-Temporal Video Semantic Role Labeling]{Constructing Holistic Spatio-Temporal Scene Graph\\ for Video Semantic Role Labeling}

\author{Yu Zhao}
\affiliation{
    \institution{College of Intelligence and Computing, Tianjin University}
    \country{}
}
\email{zhaoyucs@tju.edu.cn}

\author{Hao Fei$^*$}
\affiliation{
    \institution{NExT Research Center, National University of Singapore}
    \country{}
}
\email{haofei37@nus.edu.sg}
\thanks{* Corresponding author: Hao Fei}

\author{Yixin Cao}
\affiliation{
    \institution{Singapore Management University}
    \country{}
}
\email{caoyixin2011@gmail.com}

\author{Bobo Li}
\affiliation{
    \institution{Wuhan University}
    \country{}
}
\email{boboli@whu.edu.cn}

\author{Meishan Zhang}
\affiliation{
    \institution{Harbin Institute of Technology (Shenzhen)}
    \country{}
}
\email{zhangmeishan@hit.edu.cn}

\author{Jianguo Wei}
\affiliation{
    \institution{College of Intelligence and Computing, Tianjin University}
    \country{}
}
\email{jianguowei@tju.edu.cn}

\author{Min Zhang}
\affiliation{
    \institution{Harbin Institute of Technology (Shenzhen)}
    \country{}
}
\email{zhangmin2021@hit.edu.cn}

\author{Tat-Seng Chua}
\affiliation{
    \institution{NExT Research Center, National University of Singapore}
    \country{}
}
\email{dcscts@nus.edu.sg}

\renewcommand{\shortauthors} {Yu Zhao et al.}

\begin{abstract}
Video Semantic Role Labeling (VidSRL) aims to detect the salient events from given videos, by recognizing the predict-argument event structures and the interrelationships between events.
While recent endeavors have put forth methods for VidSRL, they can be mostly subject to two key drawbacks, including the \emph{lack of fine-grained spatial scene perception} and the \emph{insufficiently modeling of video temporality}.
Towards this end, this work explores a novel \textbf{holistic spatio-temporal scene graph} (namely HostSG) representation based on the existing dynamic scene graph structures, which well model both the fine-grained spatial semantics and temporal dynamics of videos for VidSRL.
Built upon the HostSG, we present a nichetargeting VidSRL framework.
A scene-event mapping mechanism is first designed to bridge the gap between the underlying scene structure and the high-level event semantic structure, resulting in an overall hierarchical scene-event (termed ICE) graph structure.
We further perform iterative structure refinement to optimize the ICE graph, 
such that the overall structure representation can best coincide with end task demand.
Finally, three subtask predictions of VidSRL are jointly decoded, where the end-to-end paradigm effectively avoids error propagation.
On the benchmark dataset, our framework boosts significantly over the current best-performing model. 
Further analyses are shown for a better understanding of the advances of our methods.
\end{abstract}


% \ccsdesc[500]{Computing methodologies~Natural language processing}



\maketitle

\begin{figure}[!t]
\centering
\includegraphics[width=0.98\linewidth]{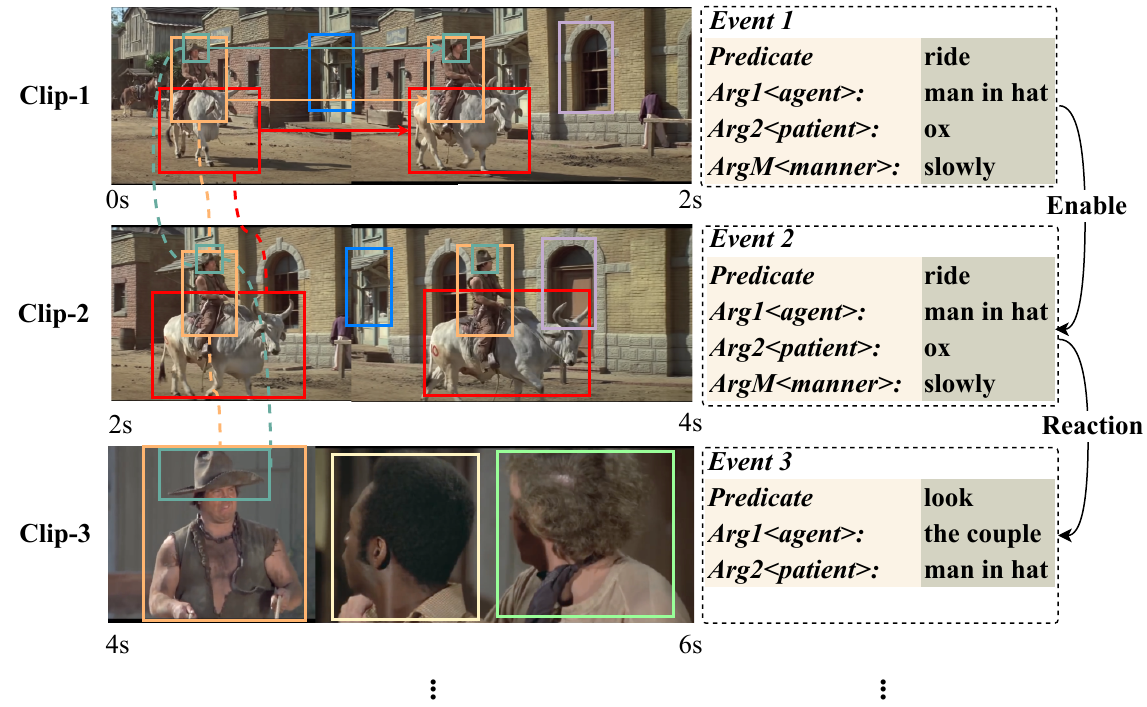}
\vspace{-3mm}
\caption{
VidSRL task cuts a video into several continuous clips (in 2s), where each clip entails one event, and each event is described by a {\it predicate} and several associated {\it arguments}.
An event also connects to another event in a certain relation.
}
\label{fig:intro}
\vspace{-4mm}
\end{figure}
\section{Introduction}
Understanding video content has become one of the foundations to human-level artificial intelligence, and further facilitating a broad range of applications, such as context recommendation \cite{DBLP:conf/mm/CaiQFHX22}, video captioning \cite{DBLP:conf/mm/NieQM00B22}, retrieval \cite{DBLP:conf/mm/CaoZWNHQ20} and robotics \cite{DBLP:journals/ijsr/KangH23, wang2023entropy, wang2023sar, wang2022recognition}. 
Unlike the spatial-level understanding of static images, videos organize a sequence of images in a temporal structure, wherein a thorough understanding of semantics relies both on the spatiality and temporality.
% along the temporality becomes more challenging.
% requires
As one of the key representative video understanding tasks, Video Role Labeling (VidSRL)\footnote{Also known as Video Situation Recognition, or Video Event Extraction.} is introduced targeting at determining the underlying video semantic structure, i.e., ``\emph{who does what to whom, where and when and how}'' within a given video \cite{DBLP:conf/cvpr/SadhuGYNK21}.
VidSRL can be broken down into three subtasks: {\bf verb prediction}, {\bf arguments generation (or semantic role labeling)} and {\bf event relation prediction}. 
Fig. \ref{fig:intro} illustrates the task with real examples.

Certain efforts have been paid to parse the semantic structure from dynamic videos correctly \cite{DBLP:journals/corr/abs-2304-00733}.
Unfortunately, existing research may largely overlook two key characteristics of the VidSRL task.
% which account for the major performance bottleneck.
% two key limitations can be found in , which 
The first one is the \textbf{lack of fine-grained spatial scene perception}.
Current methods mostly strive to extract the overall visual features of each video frame at a coarse-grained level 
% e.g., via spatial convolutions or attentions 
\cite{DBLP:conf/cvpr/SadhuGYNK21}.
Yet we notice that the event arguments always correspond to the certain fine-grained vision objects, e.g., \emph{a man in a hat} (agent) \emph{rides} (predicate) \emph{an ox} (patient) as in shown Fig. \ref{fig:intro}, where both the mentioned agent and patient are the specific objects.
% rather than the overall visual representation.
% describing % an event always 
Without a delicate grasp of spatial scenes, the semantics understanding can be hampered.
\textbf{Insufficient modeling of video temporality} is the second and key issue.
% Capturing the temporal dynamics is t
The pivot of rigorous event parsing of videos lies in precisely capturing the temporal dynamics.
Intuitively, the perception of the evolutionary movement of actions, e.g., \emph{the ox is walking forward}, is fully based on the changing of the video frames.
However, not enough attention has been paid to the temporality modeling in existing works \cite{DBLP:journals/corr/abs-2210-10828}.
Besides, existing VidSRL studies mostly ignore the interrelations between different events (clips),\footnote{Existing works ignore the task of event relation detection, and assume no interrelations between events.} which also leads to non-negligible temporal causality feature loss, as arguments are often co-referred across different events, e.g., the \emph{man} simultaneously exists in event-2 and event-3 (cf. Fig. \ref{fig:intro}).

\begin{figure}[!t]
    \centering
    \vspace{-1em}
    \hspace*{-2em}
    \subfloat{%
        \includegraphics[width=0.98\linewidth]{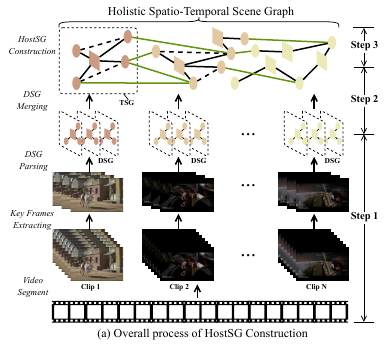}
    }
    \vspace{-1em}
    \quad
    \subfloat{%
        % \advance\leftskip+4cm
        % \hspace*{2em}
        \includegraphics[width=0.85\linewidth]{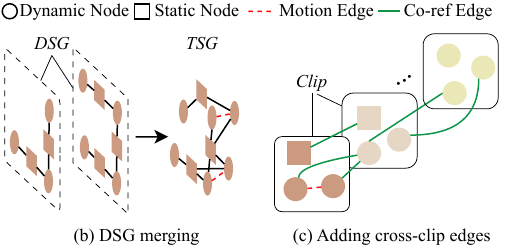}
    }
    \vspace{-2mm}
    \caption{
    Holistic spatio-temporal scene graph generation.}
    \label{fig:hostsg}
    \vspace{-6mm}
\end{figure}

To address the above challenges, in this paper we propose to construct a \emph{\textbf{ho}listic \textbf{s}patio-\textbf{t}emporal \textbf{s}cene \textbf{g}raph} (namely HostSG) for better VidSRL modeling.
% modeling the video 
We first consider modeling the video with scene graph representations \cite{DBLP:conf/cvpr/XuZCF17}.
We draw the main inspirations from the existing video dynamic scene graphs (DSGs) \cite{DBLP:journals/corr/abs-2112-09828}, where a DSG represents a video with a sequence of frame-static scene graphs.
% As frequently shown in previous studies, t
The scene graphs organize the complex visual scenes into structured graphs, where the fine-grained representations effectively depict the core semantics while filtering the less-informative background features \cite{DBLP:journals/pami/YangZC22}, which coincides with our demands.
% advance in effectively representing the visual scene semantics 
% Such structured graph representation advances in fine-grained scene semantic representation and meanwhile background noise filtering; temporal modeling ...
% Technically, we first represent each short clip with one spatio-temporal scene graph
% , which is a temporally compressed DSG for a clip.
While representing all frames with a scene graph would cause huge computational costs, we instead formulate a video clip with one temporalized DSG (TSG).
% prohibitively
TSG merges the scene graphs of all frames in one clip  into a unified graph by merging those static nodes, retaining all the spatial features meanwhile effectively reducing redundant temporal information.\footnote{Essentially, temporally compressing the DSG for a short (2s) clip would not lead to any salient information loss.}
% but greatly relieve the computation burden.
% \label{fig:hostsg}
Afterwards, the TSGs of all clips are connected into an overall holistic scene graph (i.e., HostSG) by creating the coreference edges between cross-clip nodes of DSGs, as shown in Fig. \ref{fig:hostsg}.
With the HostSG, we achieve fine-grained representations on both spatial semantics and temporal dynamics.

Built upon the HostSG, we then propose a nichetargeting VidSRL framework that performs total of five tiers of propagation (cf. Fig. \ref{fig:model}).
% , the system comprises four-tiers.
\textbf{First}, the HostSG of the input video is initialized by a graph attention model (GAT) \cite{DBLP:journals/corr/abs-1710-10903} (cf. $\S$\ref{HostMap encoding}).
% , where further the 
\textbf{Second}, we expand the HostSG by dynamically generating new event nodes. 
Then we map the underlying scene structure to the high-level event semantic structure, by 
1) creating edges between the object nodes in HostSG and the argument nodes in predicate-argument structure, 
and 2), connecting all the events along temporal dimensions,
which results in an overall \emph{h\textbf{i}erarchical s\textbf{c}ene-\textbf{e}vent} graph (termed ICE) (cf. $\S$\ref{Scene-Event Mapping}).
Intuitively, the event actions of interest could only touch on part of visual objects in a scene; also some event arguments can associate with multiple visual objects simultaneously.
Via such scene-event mapping, we bridge the gap between these two structures.
% the SG-based HostMap structure is mapped to the event semantic structure, resulting in an overall 
% The resulting hierarchical scene-event (termed ICE) graph (cf. $\S$\ref{Scene-Event Mapping}) helps bridge the gap between these two structures.
% Technically, dynamically build the ,connect each argument node
% Taking the ``Event 1'' as an example in Figure \ref{fig:intro}, the agent argument ``man in hat'' is related to two objects ``man'' and ``hat'', and the abstract modifier argument ``slow'' could not even be linked explicitly to any object.
\textbf{Third}, we perform the spatio-temporal propagation over the ICE graph, where a multipath-GAT encodes each TSG separately for the event spatial modeling, and a GRNN captures the overall event evolution and interrelations of the event-level structures along the temporal dimension (cf. $\S$\ref{Spatio-Temporal Propagation}).
\textbf{Fourth}, we further propose iteratively refining the overall ICE structure via the Graph Information Bottleneck (GIB) technique \cite{wu2020graph}, dynamically adjusting the object-argument mapping edges and the coreference edges such that the overall graph representation aligns with the end-task prediction and meanwhile screening the noisy structures (cf. $\S$\ref{Iterative Structural Refinement}).
% after a to highlight the key information and reduce noisy information.
\textbf{Fifth}, after several iterations of structural refinement, we finally perform joint predictions for the three subtasks of VidSRL based on the well-updated ICE graph (cf. $\S$\ref{Joint Task Decoding}).
Our end-to-end decoding scheme effectively avoids the error propagation introduced in the existing pipeline methods.

We perform experiments on the benchmark VidSRL data \cite{DBLP:conf/cvpr/SadhuGYNK21}, where the results show that our framework significantly boosts the current state-of-the-art (SoTA) with 0.66\textasciitilde 5.22 scores on several metrics. 
Further analyses demonstrate that the HostSG is effective in depicting the fine-grained spatial and temporal characteristics of video, and also the scene-event mapping mechanism is useful for bridging scene graph and event structure.
We also reveal how the proposed iterative refinement optimizes the ICE structure and helps lead to better predictions.

% We note that the proposed HostSG structure can be instructive for a broad range of video modeling tasks, more than VidSRL.
% Overall, this work contributes in multiple aspects:
% \begin{itemize}
%     \item We for the first time investigate modeling the fine-grained spatial semantics and temporal dynamics of videos for the VidSRL task by constructing a holistic spatio-temporal scene graph representation.
    
%     % propose improving the VidSRL task with a holistic spatio-temporal scene graph representation, which effectively models the fine-grained spatial semantics and temporal dynamics of videos.
        
%     \item We devise a scene-event mapping mechanism, bridging the gap between the underlying scene structure and the high-level event semantic structure.
    
%     \item We introduce an iterative structure refinement strategy to ensure that the resulting graph representations are best coincident with VidSRL reasoning.
    
%     \item We present an end-to-end predicting paradigm of three VidSRL subtasks, effectively voiding error propagation issue.
    
%     \item We empirically push the current SoTA VidSRL performances significantly.
%     Our codes will be made open.
%     % Anonymized codes are here: \url{}
    
% \end{itemize}

\section{Related Work}

Semantic role labeling (SRL) is a fundamental task for revealing the underlying event semantics of given data.
SRL is first pioneered in natural language processing (NLP) community, extracting the predicate-argument structure from given texts \cite{DBLP:conf/acl/GildeaJ00,DBLP:conf/emnlp/XiaLZ19,FeiGraphSynAAAI21,FeiTransiAAAI21,FeiWRLJ21}.
SRL also shares much similarity in task definition with event extraction \cite{DBLP:journals/tois/WanWXHL23,DBLP:conf/www/WuLWH23} and situation recognition \cite{DBLP:conf/aaai/YaoZSHM20,DBLP:conf/mm/ChengDLM022}.
Subsequently, the research of SRL is extended from text to vision modality, predicting the possible action labels and the corresponding argument expressions from image inputs \cite{DBLP:conf/aaai/Wei00YC22,DBLP:conf/cvpr/ChoYK22}.
Compared with text-based SRL, visual SRL (or image situation recognition) is more challenging in requiring visual spatial semantics understanding \cite{DBLP:journals/corr/GuptaM15,DBLP:conf/cvpr/YatskarZF16}.
% triplets
Later, cross-modal SRL is introduced, with both the texts and images as the input target for extracting the core predicate-argument structures \cite{DBLP:conf/mm/ZhangWZLEHLJC17,DBLP:conf/interspeech/ChibaH21}.
Multiple sources in different modalities intuitively provide complementary features and perspectives for better task understanding and prediction.

\nocite{ji2021improving,ji2022knowing,luo2021dual,wang2023towards}

Recently, the SRL has been moved to the scenario of video modality, i.e., determining the semantic role and predicate structures given video \cite{DBLP:conf/cvpr/SadhuGYNK21}.
Essentially, video is naturally suitable for describing dynamic events, due to its characteristic of sequences of video frames.
On the other hand, video understanding also raises challenges.
Different from the static scenes in image modeling, video understanding is concerned with understanding both the spatial semantics and the temporal changes \cite{DBLP:conf/iccv/Feichtenhofer0M19,DBLP:conf/cvpr/DaiYMZLLSQ22,DBLP:conf/cvpr/GengLDZ22,DBLP:conf/cvpr/WangBTT22}.
Due to the difficulty of video modeling, VidSRL has received limited research attention \cite{DBLP:conf/cvpr/SadhuGYNK21,DBLP:journals/corr/abs-2210-10828,DBLP:conf/aaai/abs-2211-01781}.
For example, Sadhu et al. (2021) \cite{DBLP:conf/cvpr/SadhuGYNK21} pioneer the task of VidSRL by encoding the video with the vision Transformer model, where the coarse-grained visual understanding leads to significant spatial semantics loss. 
Khan et al. (2022) \cite{DBLP:journals/corr/abs-2210-10828} model the VidSRL by further grounding the visual objects and entities across verb-role pairs.
Yet, they still fail to consider the video temporality modeling.
Very recently, Yang et al. (2022) \cite{DBLP:conf/aaai/abs-2211-01781} propose tracking the object-level visual arguments so as to model the changes of states.
In this regard, our work shares a similar spirit to \cite{DBLP:conf/aaai/abs-2211-01781}, but we differentiate theirs in three key aspects.
First of all, we consider a more holistic and fine-grained video representation with the video scene graph structure, in which the fine-grained objects and their correlations, as well as the temporal dynamics all can be precisely modeled.
Besides, our task formulation also considers the event relation subtask, sufficiently making use of the cross-event correlation features for better long-range dependence modeling.
Third, we consider an end-to-end task paradigm, effectively reliving the error propagation in traditional pipeline modeling.
In contrast, \cite{DBLP:conf/aaai/abs-2211-01781} ignores the event relation prediction and also takes a cascade prediction of predicates and arguments.

This work also closely relates to the topic of scene graphs (SGs).
An SG describes a given visual scene with a graph structure, where the fine-grained object nodes are connected with each other under certain relationships \cite{DBLP:conf/cvpr/XuZCF17}.
Due to the semantic-level structured representation, SGs have been widely utilized for many downstream applications \cite{DBLP:journals/mms/LuG23,DBLP:conf/cvpr/ChenJWW20}.
In this work, we consider the video SG, a.k.a., dynamic SG (DSG) \cite{DBLP:conf/dagm/HauriletAS19,DBLP:conf/cvpr/LiYX22,DBLP:conf/aaai/GengGCHRZLC21}, or spatio-temporal SG \cite{DBLP:conf/iccv/CongLARY21,DBLP:conf/cvpr/JiK0N20}, which is defined as representing each video frame with a static visual SG.
However, with the raw video SG, maintaining an SG for all the frames can be much resource-consuming, and also results in redundant SG representations.
Instead, we retrofit the video SG by merging the clip-level DSG into one single unified SG, where such renewal can effectively relieve the computational costs.

\section{Task Formulation}

Given a Video $V$ with $N$ clips $\{c_i\}_{i=1}^N$, where each clips has the 2 seconds length without overlap and describes an event.
VidSRL requires outputting $N$ related salient events $\{E_i\}_{i=1}^N$ corresponding to $N$ clips.
Each event can be represented as:
% a semantic structure:
\begin{equation}
\begin{split}
e=\left\{v,\,\langle arg^0, a^0\rangle,\,\langle arg^1, a^1\rangle,\, ...\right\},
\end{split}
\end{equation}
where $v \in \mathcal{V}$ denotes the salient event action, chosen from a pre-defined verb set $\mathcal{V}$.
$arg_j$ denotes the argument roles of $v$, each of which is specified with a free text span $a_j$.
Each predicate has up to 10 arguments which is also predefined by VidSRL \cite{DBLP:conf/cvpr/SadhuGYNK21}.

\section{Holistic Spatio-temporal Scene Graph Construction}

As cast earlier, we propose to construct a holistic spatio-temporal scene graph (HostSG) to model fine-grained object-level spatial and temporal characteristics of videos.
The construction of HostSG is a three-step process, as shown in Fig. \ref{fig:hostsg}(a).
First, we generate DSG for each video clip, containing several scene graphs.
Then we merge each DSG (a set of static scene graphs) of a clip into an integrated graph, i.e., the temporalized DSG (TSG).
Finally, we combine TSGs of all clips (each clip has one TSG) into the overall HostSG by adding cross-clip coreference edges.

\paragraph{\bf Step-1: DSG Generation for Clip}
We suppose the videos are already segmented into clips without overlap in timeline.
We first perform keyframe extraction for each raw clip, so as to maintain the key semantics meanwhile reducing redundant features.
To exactly capture the salient event content of the video clip and to avoid producing overlapping scene graphs from similar or blurry frames, we adopt a clustering-based extraction method \cite{song2016click} to extract significant keyframes from the raw video clips. 
Then, these frames are fed into the scene graph parser MOTIFS-TDE \cite{DBLP:conf/cvpr/TangNHSZ20} for producing the $DSG=\{G_t\}$, consisting of a series of static scene graph $G_t=(V_t,E_t)$ of each keyframe of time $t\in T_c$ for clip $c$.
The scene graph parser outputs a probability score for each node, and only the top-$k$ ones will be selected for each $G_t$.

\begin{figure*}[!t]
    \centering
    \includegraphics[width=0.98\linewidth]{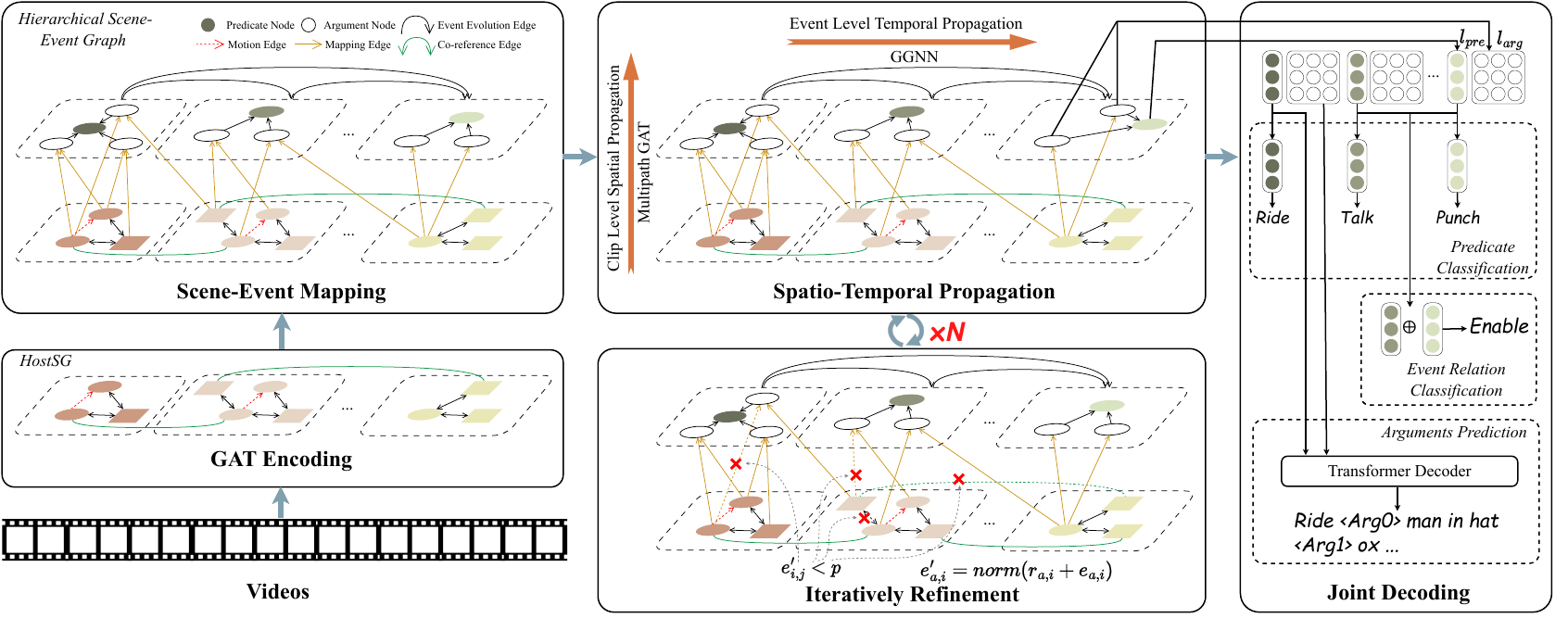}
    \vspace{-2mm}
    \caption{
    Augmented holistic event-arguments semantic graph.}
    \label{fig:model}
    \vspace{-2mm}
\end{figure*}

\paragraph{\bf Step-2: Merging DSG}
Although we select the keyframes from the raw video clip, the resulting DSG of each clip can still contain many redundant nodes and edges that are unnecessary for computation and propagation \cite{wu2020graph}.
Also the DSG contains separate SG in sequence, where the temporal correlations are not explicitly modeled.
Inspired by \cite{cherian20222}, we combine the DSG of a clip into one temporalized Dynamic Scene Graph (TSG), representing both the spatial appearance information and the temporal object motions.

Concretely, according to the labels of the DSG object nodes, 
we divide all the objects into static and dynamic ones.
The static nodes (e.g., house, street, trees) refer to the background objects that may not move observably within a short time, and the dynamic nodes (e.g., person, car, animal) are the opposite.
We merge those static nodes across timeline based on their coreference, i.e., with highly identical categories and locations.
Two static nodes $v_t \in G_t,v_{t'}\in G_{t'} (t\neq t')$ will be merged if they meet:
\begin{equation}
    tag_{v_t}=tag_{v_{t'}}\quad \text{and} \quad IoU(loc_{v_t},loc_{v_{t'}})>d,
\end{equation}
where $tag$ and $loc$ represent the label and bounding box of an object, and $IoU$ means the intersection over the union of two boxes.
If a node has multiple candidate nodes in a frame, the one with the nearest centroid will be selected.
After merging, we keep their original edges with other nodes.
For dynamic nodes, instead of merging, we take the same way to recognize the coreference nodes but add edges between them to represent the motion relation.
Then the DSG $\{G_t\}$ is compressed to a TSG $G^{tsg}_c=(V^{tsg}_c,E^{tsg}_c)$ for clip $c$.
The process is illustrated in Fig. \ref{fig:hostsg}(b).

\paragraph{\bf Step-3: HostSG Construction}
While a TSG depicts the inner-clip information, the cross-clip relations are still not well represented.
To co-view all the clips holistically, we now combine the TSGs into one unified graph, by adding coreference edges between object nodes.
However, due to the significant time interval between frames of different clips, there may be shot transitions or large movement of objects, which largely makes the IoU lose effectiveness.
Here we instead link all TSGs
by considering the semantic similarity between node visual features as well as their labels:
% We calculate the semantic similarity by:
\begin{equation}
    sim(v_s,v_o)= \text{Softmax} (Relu(h_s\cdot h_o)),
\end{equation}\label{eq:3}
where $h_s,h_s$ denote the visual features of node $v_s\in V^{tsg}_c,v_o\in V^{tsg}_{c'} (c\neq c')$, respectively.
% , obtained by the scene graph generator.
Then we link the nodes meeting the following condition:
\begin{equation}
    tag_{v_s}=tag_{v_o}\quad and \quad sim(v_s,v_o) > \gamma,
\end{equation}
where $v_s$ and $v_o$ are from different TSGs.
If a node has multiple candidate nodes in another frame to link, the candidate with maximum $sim$ is selected:
\begin{equation}
    match(v_s)=\mathop{argmax}_{v_{o'}\in \{v|tag_{v}=tag{_{v_s}}\}} (sim(v_s,v_{o'})).
\end{equation}

For dynamic nodes in a TSG, there can be multiple nodes representing one object.
To tackle this, we treat all the dynamic nodes linked by motion edges as a whole, and average their representations for the similarity calculation:
\begin{equation}
    sim(\bm{V}_s,\bm{V}_o)=\text{Softmax}(Relu(Avg(\bm{H}_d)\cdot Avg(\bm{H}_o))),
\end{equation}
where $\bm{V}_s$ and $\bm{V}_o$ are sets of dynamic nodes linked with motion edges.
Once two sets of dynamic nodes are matched, we combine them by linking a node in one set with all the nodes in the other sets, producing a coreference cluster.

By traversing all the clip pairs with the above process, we can finally obtain the HostSG for the overall video.
Here we denote the HostSG of a video as $\{G|(V,E)\}$, where $V$ is the node set containing static nodes and dynamic nodes from all the TSGs, and $E$ is the edge set containing scene graph edges, motion edges and coreference edges.
We provide more details of the HostSG construction in Appendix $\S$C.

\section{VidSRL Framework}

We propose a hierarchical framework to perform the VidSRL task.
As shown in Fig. \ref{fig:model}, the overall framework consists of five tiers: HostSG encoding, scene-event mapping, spatial-temporal propagation, iterative structure refinement and joint decoding.

\subsection{HostSG Encoding}
\label{HostMap encoding}
Given a HostSG $G$ of $N$ video clips, 
we denote the node representations of $G$ as $\bm{X}=\{x_1,x_2,...,x_m\}$, where $x_i$ are initialized by the concatenation of the object visual features and the corresponding node label embeddings, both of which are obtained from a pre-trained scene graph parser, and a pre-trained language model encoder, respectively.
Then a graph attention network (GAT) is used to encode $G$:
\begin{equation}\label{eq:7}
    \bm{X}_{host}=\mathop{\text{GAT}}_{\theta_{base}}(\bm{X}).
\end{equation}

\subsection{Scene-Event Mapping}
\label{Scene-Event Mapping}

An event structure consists of a predicate and several arguments, where the predicate reflects the event action and the arguments describe the related roles or modifiers.
For instance, the structure of event-1 as in Fig. \ref{fig:intro} contains a predicate ``ride'', and three arguments, ``man in hat'', ``ox'' and ``slowly''.
Such high-level event semantic structure does not well align with the low-level scene-based HostSG structure.
Thus, in this step we aim to bridge the gap between these two structures.
% Mapping SG and Event structure into scene-event semantic graph: SESG.
% And then encode SESG.
% % \paragraph{\bf SESG construction via Mapping}
% Intuitively, the raw TSG with the object-relation structure could not represent an event-argument structure (Figure1 \ref{}).
% Hence during this step
Technically, we connect the scene nodes of HostTSG and the event nodes of predicate-argument structure, leading to a broader structure of hierarchical scene-event graph (ICE).
% adopt a semantic mapping mechanism to bridge the gap between scene graph structure and event structure, extending the HostTSG to a hierarchical scene-event graph (ICE).
% Concretely

As shown in Fig. \ref{fig:model}, for each TSG $G^{tsg}$ in HostSG $G^{tsg}\subset G$, we dynamically maintain a corresponding event structure $G^{evt}$ where the predicate node connects to all the argument nodes.
% generate certain abstract nodes to represent the event predicate and arguments.
In accordance with the VidSRL definition \cite{DBLP:conf/cvpr/SadhuGYNK21}, we maintain 11 event nodes, including a predicate node, 5 real argument nodes, and 5 modifier argument nodes.
% , according to its definition.
% The predefined arguments and their meanings are list in Table \ref{tab:args}.
% \input{tables/arguments}  appendix=============
Then, for each argument node $v_e$ in $G^{evt}$, we tentatively link all object nodes in the corresponding TSG $G^{tsg}$ to $v_e$, which will be further refined through training.
Further, we create the edges between predicate nodes across all events, in which each predicate node directionally points to all the subsequent ones.
Such cross-event connections help capture the event evolutionary relationships.
The predicate node is initialized by averaging all the node representations of the TSG; while the argument nodes are initialized with the 10 argument embeddings:
\begin{equation}\label{eq:8}
    \bm{X}_{evt}=\{\bm{x}_{p}=\text{Avg} \, (V^{tsg}),\bm{x}_{arg_i}=\text{Embed}(Arg_i),i\in[1,10]\}.
\end{equation}

% With the added nodes and edges, we obtain the ICE, whose whole structure is illustrated in Figure \ref{fig:model}.

\subsection{Spatio-Temporal Propagation}
\label{Spatio-Temporal Propagation}

With the overall ICE graph, we now perform spatio-temporal propagation for a thorough feature learning.
The message passing is divided into the spatial one at the clip level and the temporal one at the event level, subsequently.

\paragraph{\bf Clip-level Spatial Propagation with Multipath-GAT}
% Within a clip, w
We encode the ICE graph at each clip dimension, which is the combination of the $G^{tsg}$ and $G^{evt}$ at the corresponding clip.
Since we define multiple types of nodes and edges to represent various semantics, we consider a multipath-GAT for encoding, which keeps the information propagation separately among different types of nodes.
Technically, based on the multi-head architecture of GAT, we divide these heads into three groups: coreference group, event group and normal group.
The coreference group encodes only the object nodes of TSGs; the event group encodes only the abstract event nodes (i.e., predicate nodes and argument nodes) and the normal group encodes all the nodes:
\begin{equation}
\begin{split}
    \bm{H'}_{host}=\mathop{\text{GAT}}&_{\theta_{obj}}(\bm{X}_{host}),\quad
    \bm{H'}_{evt}=\mathop{\text{GAT}}_{\theta_{evt}}(\bm{X}_{evt}), \\
    \bm{H'}_{norm}&=\mathop{\text{GAT}}_{\theta_{norm}}([\bm{X}_{host}, \bm{X}_{evt}]), \\
\end{split}
\end{equation}
where $\bm{X}_{host}$ is the representation of object, obtained from the base scene graph encoder by Eq. (\ref{eq:7}).
$\bm{X}_{evt}$ is obtained by Eq. (\ref{eq:8}).
$\theta_{obj}$, $\theta_{evt}$ and $\theta_{norm}$ are the three groups of GAT parameters.
$\bm{H'}_{obj}$, $\bm{H'}_{evt}$ and $\bm{H'}_{norm}$ are the node representations of the final GAT layer.
Next, we combine the three types of nodes with an element-wise summation:
\begin{equation}
    \bm{H}=[\bm{H'}_{host};\bm{H'}_{evt}] + \bm{H'}_{norm}.
\end{equation}

\paragraph{\bf Event-level Temporal Propagation with GGNN}
By encoding the spatial features of each clip or event, we now process temporal information aggregation among events.
Specifically, a Gated Graph Sequence Neural Network (GGNN) \cite{DBLP:journals/corr/LiTBZ15} is adopted to propagate over all the event structures in ICE: 
\begin{equation}
    \bm{L}_{p}=\text{GGNN}(\bm{H}_{p}),
\end{equation}
where $\bm{H}_{p}$ is the representations of predicate nodes obtained from $\bm{H}$, and $\bm{L}_{p}$ is the final layer outputs of the GGNN.

\subsection{Iterative Structural Refinement}
\label{Iterative Structural Refinement}
At each end of the spatio-temporal propagation, we further perform structural refinement over the ICE structure.
% After each GAT layer, we insert a series of operations to adapt the co-reference edges and argument-object mapping edges
Previously, we build the fully-connected initial edges of argument-object mapping, while the realistic case can be that one argument may also associate with a very small part of the objects in scenes.
Also the initial structures in terms of the event evolution edges as well as the coreference edges among TSGs should also be updated to best meet the task prediction.
To this end, we iteratively adjust the overall structure.
Each refinement step is processed after one GAT layer.

\begin{table*}[t]
\fontsize{4}{5}\selectfont
\setlength{\tabcolsep}{1.3mm}
\caption{
Main results on the VidSRL dataset. ``Verb Cls'', ``SRL'' and ``EvtRel'' represents the three subtasks verb classification, semantic role labeling and event relation prediction.  The CIDEr score is also computed over every verb-sense (CIDEr-Verb) and over argument-types (CIDEr-Arg).
Bold numbers are the best, and underlined ones are the second best.
Our results are averaged on five running with different seeds.
Gray color: methods use ground-truth verb annotations for SRL training.
% \scott{What's grey color?}
}
\vspace{-2mm}
\begin{center}
\resizebox{0.98\textwidth}{!}{
\begin{tabular}{clccccccccccccc}
\hline
\multicolumn{2}{c}{\multirow{2}{*}{}} & \multicolumn{3}{c}{\bf \texttt{Verb Cls}} & \multicolumn{6}{c}{\bf \texttt{SRL}} & \multicolumn{1}{c}{\bf \texttt{EvRel}} \\
\cmidrule(r){3-5} \cmidrule(r){6-11} \cmidrule(r){12-12}
\multicolumn{2}{c}{} &\bf Acc@1(\%) & \bf Acc@5(\%) &\bf Rec@5(\%) &\bf CIDEr &\bf Rouge-L &\bf CIDEr-Vb &\bf CIDEr-Arg &\bf Lea &\bf Lea-S &\bf Macro-Acc(\%) \\
\hline
\multicolumn{12}{l}{$\bullet$ \textbf{Pipeline}} \\
& VidSitu-GPT2 \cite{DBLP:conf/cvpr/SadhuGYNK21} & - & - & - & \cellcolor{gray!40} 34.67 & \cellcolor{gray!40} 40.08 & \cellcolor{gray!40} 42.97 & \cellcolor{gray!40} 34.45 & \cellcolor{gray!40} 48.08 & \cellcolor{gray!40} 28.10 & - \\
& VidSitu-I3D \cite{DBLP:conf/cvpr/SadhuGYNK21} & 30.17 & 66.83 & 4.88 & \cellcolor{gray!40} 47.06 & \cellcolor{gray!40} 42.41 & \cellcolor{gray!40} 51.67 & \cellcolor{gray!40} 42.76 & \cellcolor{gray!40} 48.92 & \cellcolor{gray!40} 33.58 & - \\  
& VidSitu-SlowFast \cite{DBLP:conf/cvpr/SadhuGYNK21} & 32.64 & 69.20 & 6.11 & \cellcolor{gray!40} 45.52 & \cellcolor{gray!40} 42.66 & \cellcolor{gray!40} 55.47 & \cellcolor{gray!40} 42.82 & \cellcolor{gray!40} \underline{50.48} & \cellcolor{gray!40} 31.99 & \underline{34.13} \\  
% & TxDec & 53.59 & 41.74 & 77.68 & 469.23 & 67.03 & \underline{22.53} & \underline{24.93} & 51.27 & 227.29 & 41.63 \\
\hline
\multicolumn{12}{l}{$\bullet$ \textbf{Joint}}\\
& VidSitu-e2e \cite{DBLP:conf/aaai/abs-2211-01781} & 46.79 & 75.90 & 23.38 & 30.33 & 29.98 & 39.56 & 23.97 & 35.92 & - & - \\
& OME \cite{DBLP:conf/aaai/abs-2211-01781} & 52.75 & 83.88 & 28.44 & 47.82 & 40.91 & 54.51 & 44.32 & - & - & - \\
& OME(disp) \cite{DBLP:conf/aaai/abs-2211-01781} & 53.32 & \underline{84.00} & 28.61 & 48.46 & \underline{41.89} & 56.04 & \underline{44.60} & - & - & - \\
& OME(disp)+OIE \cite{DBLP:conf/aaai/abs-2211-01781} & \underline{53.36} & 83.94 & \underline{28.72} & 47.16 & 40.86 & 53.96 & 42.78 & - & - & - \\
& VideoWhisperer \cite{DBLP:journals/corr/abs-2210-10828} & 45.06 & 75.59 & 25.25 & \underline{52.30} & 35.84 & \underline{61.77} & 38.18 & 38.00 & - & - \\
\hline
% \multicolumn{12}{l}{$\bullet$ \textbf{VL-PTMs + 3D scene features}} \\
& \bf HostSG (Ours) & \textbf{56.15} & \textbf{86.33} & \textbf{29.38} & \textbf{55.09} & \textbf{43.13} & \textbf{64.24} & \textbf{47.68} & \textbf{55.70} & \textbf{35.01} & \textbf{35.97} \\
\specialrule{0em}{-2pt}{-1pt}& & \fontsize{3}{5.5}\selectfont{(+2.79})  & \fontsize{3}{3.5}\selectfont{(+2.33})  & \fontsize{3}{1.26}\selectfont{(+0.66}) & \fontsize{3}{3.5}\selectfont{(+2.79})  & \fontsize{3}{3.5}\selectfont{(+1.24})  & \fontsize{3}{2.47}\selectfont{(+2.47}) & \fontsize{3}{3.5}\selectfont{(+3.08})  & \fontsize{3}{3.5}\selectfont{(+5.22})  & \fontsize{3}{3.5}\selectfont{(+3.2})  & \fontsize{3}{3.5}\selectfont{(+1.84}) \\
\hline
\end{tabular}
}
\label{tab:result}
\end{center}
\vspace{-3mm}
\end{table*}

\paragraph{\bf Edge weights adjustment}
We refine the graph structure via adjusting the edge weights.
The ICE has five types of edge, i.e., DSG edges, coreference edges, scene-event mapping edges, predicate-argument edges and event evolution edges.
The predicate-argument edges are fixed due to the inherent event structure, and we adjust the other four types.
We first calculate a relation score $r_{a,i}$ between a pair of nodes $a$ and $i$:
% For coreference edges, we re-calculate the similarity score by Equation \ref{eq:3} and add to the current edge weights:
% \begin{equation}
%     e'_{i,j}=e_{i,j}+sim(v_i,v_j)
% \end{equation}
% Then, we filter the coreference edges (set the weight to 0) with a threshold $d$ when $e'_{i,j}<d$.
% For coreference clusters, we travel each TSG node through coreference edges and motion edges to get its k-order neighbours, and average the weights of all the coreference edges among them. 
% At last, we normalize the weights among all the coreference edges. 
% With the above steps, we could accumulate the edge weights in a real coreference cluster and dilute the weights of the other edges.

% \paragraph{\bf Argument-object mapping edge adjustment}
% For argument-object mapping edges, we refine their weights with an argument-object relation score:
\begin{equation}
\begin{split}
    \bm{a}_{a,i}&=\frac{\exp{(\bm{W_1}[\bm{h}_a;\bm{h}_i])}}{\sum_{j\in \Phi}\exp{(\bm{W_1}[\bm{h}_a;\bm{h}_j])}}, \\
r_{a,i}&= Softmax(FFN(\bm{a}_{a,i})),
\end{split}
\end{equation}
where $\bm{h}_a$ and $\bm{h}_i$ are the hidden representations of node $a$ and $i$, $\Phi$ denotes the set of all nodes in ICE.
$\bm{W_1}$ and $\bm{W_2}$ are learnable parameters.
% Besides, we also take into account the label of object nodes, and the final edge weight is:
% \begin{equation}
    % e_{a,i}=\text{norm}(r_{a,i}\cdot\cos{(\bm{h}_a,\bm{t}_i)})
% \end{equation}
% where $\cos{(\cdot)}$ is cosine distance, $\bm{t}_a$ is the label embedding of the object node $i$.
Then we change the weight of edge $e_{a,i}$ to:
\begin{equation}
    e'_{a,i}=\text{norm}(r_{a,i}\cdot e_{a,i}).
\end{equation}
Afterwards, we filter the edges with low $e'_{a,i}$ by setting its weight to 0.
Specifically, we further cut the coreference edges between nodes with different labels due to the natural constraint.
Through edge adjustment, the model will adaptively filter noisy edges (setting weights to 0) and meanwhile newly add informative edges (setting weights to non-zero).

\paragraph{\bf Assistant optimization}
Thereafter, we obtain an adjusted ICE from the final layer refinement, i.e., $G^{-}$, and its node representations $\bm{H}^-$.
We perform pooling over $\bm{H}$ to obtain the graph representation $\bm{g}$ and further concatenate it with predicate nodes as the context features for event verb and event pair relation:
\begin{equation}
\begin{split}
    \bm{a}_v=[\bm{g};\bm{h}^-],\quad
    \bm{a}_{rel}=[\bm{g};\bm{h}^-_i;\bm{h}^-_j],
\end{split}
\end{equation}
where $\bm{h}\in \bm{H^-}$ is the representation of the predicate node.

To ensure the adjusted coreference edges and mapping edges are informative enough, we consider an assistant optimization objectives.
Inspired by \cite{wu2020graph}, we denote $\bm{z}_a, \bm{z}_{rel}$ as the compact information of the resulting $G^-$, which is sampled from a Gaussian distribution parameterized by $\bm{a}_v$ and $\bm{a}_{rel}$.
The we adopt the GIB-driven loss:
\begin{equation}
    \mathcal{L}_{GIB}=\min[-I(\bm{z}_v,Y_v)-I(\bm{z}_{rel},Y_{rel})+I(\bm{z}_v, G)+I(\bm{z}_{rel}, G)],
\end{equation}
where $I(\cdot)$ denotes the mutual information.
In appendix $\S$A, we give the detailed technical process of GIB optimization.

\subsection{Joint Task Decoding}
\label{Joint Task Decoding}
In the decoding stage, we use three heads for the three subtasks of VidSRL separately.
For verb prediction and event relation classification, we use the MLP classifiers respectively.
\begin{equation}
\begin{split}
    \bm{v}_i=\mathop{\text{MLP}}_{verb}(\bm{l}_{p_i}),\quad
    \bm{r}(i\rightarrow j)=\mathop{\text{MLP}}_{evtrel}([\bm{l}_{p_i};\bm{l}_{p_j}]),
\end{split}
\end{equation}
where $\bm{l}_{p_i},\bm{l}_{p_j}\in \bm{L}_{p}$ is the predicate node representation of event $i$.
$v_i$ and $r(i\rightarrow j)$ are further passed through a softmax function and predict the verb for event $e$ and the event relation from event $i$ to event $j$.

For argument generation, we adopt a transformer decoder to generate the arguments in a sequence-to-sequence manner.
Different from \cite{DBLP:conf/cvpr/SadhuGYNK21}, we passed argument node features as the input token embeddings:
\begin{equation}
    args=\text{TextDecoder}(\bm{l}_{p_i},\bm{h}_{arg_1},\bm{h}_{arg_2},...\bm{h}_{arg_{10}}),
\end{equation}
where $\bm{h}_{arg_i}\in\bm{H}$ is the argument node representation.

\subsection{Training Objectives}
\label{Training Objectives}
For the three subtasks of VidSRL, i.e., verb prediction, event relation prediction and arguments generation, we use cross-entropy between decoder outputs and the ground-truth as the objectives, denoted by $\mathcal{L}_{v}$, $\mathcal{L}_{r}$ and $\mathcal{L}_{arg}$.
Based on the objectives of the three subtasks and the assistant GIB loss of graph refinement, we set the final joint loss as:
\begin{equation}
\mathcal{L}=\mathcal{L}_v+\eta_1\mathcal{L}_{r}+\eta_2\mathcal{L}_{arg}+\eta_3\mathcal{L}_{GIB}.
\end{equation}

\section{Experiments}

\subsection{Settings}
% \paragraph{\bf Dataset and Evaluation}
We experiment with the VidSRL data \cite{DBLP:conf/cvpr/SadhuGYNK21}, which contains 28K videos from movies.
Each video is 10 seconds in length, and is cut into five clips with 2 seconds each.
VidSRL annotations contain 2K+ verb categories with more than three arguments (semantic roles) each.
% Table \ref{tab:result} give the dataset statistics.
For fair comparison, we follow \cite{DBLP:conf/cvpr/SadhuGYNK21} to measure verb prediction with Acc@K and Recall@K, and use top-1 accuracy for event relation classification.
For free-form argument generation, we follow \cite{DBLP:conf/cvpr/SadhuGYNK21} to measure the prediction of ARG0, ARG$_1$, ARG$_2$, ARG$_{LOC}$ and ARG$_{SCENE}$, using CIDEr, ROUGE-L to measure agreement computation and Lea \cite{DBLP:conf/acl/Moosavi016} to measure coreferencing capability.
% \paragraph{\bf Baseline Overview}
We compare our model with recent SoTA methods with different architectures.
We divide all the methods to {\bf 1) pipeline} and the {\bf 2) joint} ones, where the pipeline methods train modules for the three VidSRL tasks separately, while the joint methods train one model jointly.
The pipeline methods may use the golden-standard verb-role pairs as inputs for SRL subtask, while the joint methods train the three tasks together with only videos as inputs.
% \paragraph{\bf Implementation Details}
For DSG parsing, we extract 5 key frames from each clips, and adopt MOTIFS-TD \cite{DBLP:conf/cvpr/TangNHSZ20} to get the scene graphs and keep the top 15 objects.
We follow \cite{cherian20222} for dividing static and dynamic nodes when merging the DSGs. 
% \scott{need specify SG parsing details.}
Model hyperparameters and more detailed experiment settings are shown in Appendix $\S$B.

\subsection{Main Results}
Table \ref{tab:result} shows the overall results.
Overall, the new proposed joint models outperform the baseline pipeline methods.
For the verb classification task, we can see the joint models with far better features achieve significant improvements than the baselines with only I3D and SlowFast features.
Note that the ``VidSitu-GPT2'' line means it only uses a visual-blind language model to predict arguments with the textual inputs of verbs.
Thus it has no results for verb and event relation prediction.
``VidSitu-GPT2'' only achieves a very low performance due to the missing of video information.

For SRL subtask, the results of pipelines take advantage of the ground-truth annotations of verbs by teacher-forcing thus we mark them with the gray color.
Instead, we report the reproduced results of \cite{DBLP:journals/corr/abs-2210-10828} in ``VidSitu-e2e'' for a fair comparison.
We can see the  ``OME'' and ``VideoWhisperer'' models achieve remarked progress from ``VidSitu-e2e''.
In contrast, our proposed method achieves significantly better results over three tasks on all metrics, demonstrating its efficacy.
Furthermore, our model could cover the event relation prediction task while most other models do not.

\begin{table}[t]
\fontsize{9}{11.5}\selectfont
\setlength{\tabcolsep}{1.mm}
\begin{center}
% \resizebox{1\textwidth}{!}{
\caption{Module ablation results.
`w/o SG features' means ablating the graph modeling of scene graph feature, while instead using the overall video features.
`w/o SG-Evt mapping' means replacing the generated event nodes with the average pooling of the object nodes features.
`w/o Refinement' means ablating the iterative refinement.
% using the average-pooling results of all the nodes.
% replacing with beam search decoding.
`w/o Multipath GAT' means replacing the multipath GAT with a normal GAT.
`Base-GAT' and `GGNN' represents the base encoder and the GGNN network before decoder.
}
\vspace{-2mm}
\begin{tabular}{lccc}
\hline
\multicolumn{1}{c}{} & \bf Acc@1 & \bf CIDEr & \bf Macro-Acc \\
% \hline
% \multicolumn{4}{l}{\textbf{2D vs. 3D Info}} \\
% & 2D & - & - \\
% & 3D & - & - \\
\hline
\textsc{HostSG} (Full) & \bf 56.15 & \bf 55.09 & \bf35.97 \\
% \cdashline{1-4}
% \hline
\quad w/o SG features & 46.19(-9.96) & 47.38(-7.71) & 34.40(-1.57)\\
\quad w/o SG-Evt mapping & 53.43(-2.72) & 52.38(-2.71) & 34.81(-1.16) \\
\quad w/o Refinement & 54.51(-1.64) & 52.85(-2.24) & 35.06(-0.91) \\
\quad w/o Multipath GAT & 54.92(-1.23) & 53.35(-1.74) & 35.50(-0.47) \\
% & w/o Global Visual Feature & 23.91 & 45.70 \\
\quad w/o \textsc{Base-GAT} & 55.10(-1.05) & 54.93(-0.96) & 35.83(-0.16)\\
\quad w/o GGNN & 55.77(-0.38) & 55.01(-0.08) & 35.00(-0.97) \\

\hline
\end{tabular}
% }
\label{tab:ablation}
\end{center}
\vspace{-3mm}
\end{table}

% For ``w/o \textsc{Go3D}-S$^2$G'', we 

% removing the graph feature $r^{G}$ and feed the embedded 3D features of target objects $\bm{s}_i^v$ in Equation \ref{equation:v} to the encoder for text generation.
% For ``w/o O\textsc{C}GCN'', we replace the O\textsc{C}GCN with a normal GCN.
% For ``w/o S$^3$ mechanism'', we using the average-pooling results of all the nodes in the scene graph as graph feature for text generation.
\textbf{Model Ablation.} Now we quantify the contribution of each design in our systems via model ablation, as shown in Table \ref{tab:ablation}.
First, we can see that the scene graph features give the biggest influence, contributing 9.96 Acc for verb classification, 7.71 CIDEr for argument prediction and 1.57 for event relation classification.
Besides, the scene-event mapping and the refinement mechanism also play an essential role in our framework.
The GGNN mainly contributes to the event relation classification tasks, gaining 0.97 Acc.
Other module designs, the base GAT and multipath-GAT, enhance the system more or less.

\subsection{Analysis and Discussion}
% To gain an in-depth understanding of our methods, we further carry out detailed analyses under several different aspects.

% the effectiveness of 

% \paragraph{\noindent$\blacktriangleright$ \bf Q1: \textbf{Does HostSG provide informative spatial and temporal features for VidSRL?}}
\paratitle{$\blacktriangleright$ \bf Q1: \textbf{Does HostSG provide informative spatial and temporal features for VidSRL?}}
The primary motivation of our method is that the HostSG could provide fine-grained spatial and temporal features.
We now downgrade the HostSG with the object features without a graph structure, and replace $\bm{X}_{host}$ with the raw object features, keeping the event nodes in ICE.
As shown in Figure \ref{fig:sg_obj}, the modeling with object features decreases the performance markedly.

We also explore the temporal modeling of inner-clip DSG merging.
When extracting frames, we try 11, 5 and 1 frames for each clip separately.
When the number of frames decays to 1, there is no temporal information then.
As shown in Table \ref{tab:frames}, single frames without temporal information abate final performance, while the situations on 5 or 11 frames have consistent performance.
We suppose that for a 2s clip, five keyframes are enough to capture the salient objects from different temporal scenes.
Too many frames may bring noisy information, while single frame may lose some objects due to the potential scene transition.

\begin{table}[t]
    \fontsize{9}{11.5}\selectfont
    \setlength{\tabcolsep}{1.mm}
    \begin{center}
    % \resizebox{1\textwidth}{!}{
    \caption{Influence of different numbers of frame extraction. `w/o Key Frame Extraction' means we extract frames with a constant interval.}
    \begin{tabular}{clccc}
    \hline
    \multicolumn{2}{c}{} & \bf Acc@1 & \bf CIDEr & \bf Macro-Acc \\
    % \hline
    % \multicolumn{4}{l}{\textbf{2D vs. 3D Info}} \\
    % & 2D & - & - \\
    % & 3D & - & - \\
    \hline
    & $\bullet$ 1 Frame/Clip & 41.48 & 36.85 & 33.91 \\
    & \quad w/o Key Frame Extraction & 41.51 & 37.10 & 34.02 \\
    \hline
    & $\bullet$ 5 Frames/Clip & \bf 56.15 & \bf 55.09 & \bf 35.97 \\
    & \quad w/o Key Frame Extraction & 56.13 & 54.77 & 35.16 \\
    \hline
    & $\bullet$ 11 Frames/Clip & 55.15 & 54.72 & 35.31 \\
    & \quad w/o Key Frame Extraction & 55.04 & 54.67 & 35.29 \\

    \hline
    \end{tabular}
    % }
    % \vspace{-2mm}
    \label{tab:frames}
    \end{center}
    \vspace{-2mm}
    \end{table}

    % For ``w/o \textsc{Go3D}-S$^2$G'', we 
    
    % removing the graph feature $r^{G}$ and feed the embedded 3D features of target objects $\bm{s}_i^v$ in Equation \ref{equation:v} to the encoder for text generation.
    % For ``w/o O\textsc{C}GCN'', we replace the O\textsc{C}GCN with a normal GCN.
    % For ``w/o S$^3$ mechanism'', we using the average-pooling results of all the nodes in the scene graph as graph feature for text generation.
% \pgfplotsset{compat=1.7,every axis title/.append style={at={(0.5,-0.7)}}}
\pgfplotsset{compat=1.7,every axis title/.append style={at={(0.5,-0.45)}, font=\fontsize{14}{1}\selectfont},every axis/.append style={xtick pos=left,ytick pos=left,tickwidth=1.5pt}}
\usetikzlibrary{matrix}
\usepgfplotslibrary{groupplots}
\usetikzlibrary{patterns,backgrounds}

\definecolor{c1}{RGB}{252,232,212}
\definecolor{c2}{RGB}{184,183,163}
\definecolor{c3}{RGB}{107,112,092}
\definecolor{c4}{RGB}{203,153,126}
\definecolor{c5}{RGB}{107,112,092}

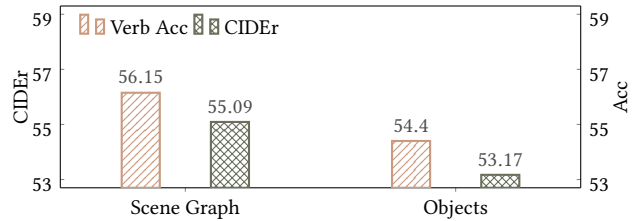
\begin{figure}[t]
\centering
\begin{tikzpicture}
%\begin{groupplot}[group style={group name=myplot,group size=2 by 1,horizontal sep=30pt,xlabels at=edge bottom, ylabels at=edge left},height=4cm,width=4cm]
\begin{axis}[
	ybar,
	ytick = {53, 55, 57,59},
	yticklabels={53, 55, 57,59},
	ymax=59,ymin=53,
	ylabel = CIDEr,
	y tick label style = {yshift=-0.5em, text height=0ex,font=\small},
	%    ytick style = {yshift=-0.0em, font=\scriptsize},
    x label style = {font=\small},
    y label style = {yshift=-0.5em, font=\small},
    axis x line*=bottom,
	axis line style={-},
    axis y line*=left,
	axis line style={-},
	enlargelimits=0.05,
    nodes near coords,
    every node near coord/.append style={black, font=\small, opacity=0.7, yshift=-0.0em,xshift=0.0em },
	legend style={at={(0.02,0.99)},anchor=north west, draw=none, legend columns=-1, font=\small},
	xticklabels={Scene Graph,Objects},
	xtick={2,6},
% 	ymajorgrids,
	xmax=7.5, xmin=0.5,
	x tick label style = {yshift=0.05em, align=center,font=\small},
%	title=(a) BLEU-4,
	title style={yshift=-0.9em,font=\small},
	width = 8.5cm,
	height = 4cm,
	]
	\addplot[c4, pattern= north east lines, thick, pattern color=c4, bar shift=0pt, bar width = 1.6em] coordinates
	{
		(1.34,  56.15)
	};\addlegendentry{Verb Acc}
	
	\addplot[c3, pattern=crosshatch, pattern color=c3, thick, bar shift=0pt, bar width = 1.6em] coordinates
	{
		(2.66,  55.09)
	};\addlegendentry{CIDEr}
\end{axis}

\begin{axis}[
	ybar,
	ytick = {53, 55, 57,59},
	yticklabels={53, 55, 57,59},
	ymax=59,ymin=53,
	ylabel = Acc,
	y tick label style = {yshift=-0.5em, text height=0ex,font=\small},
%   ytick style = {yshift=-0.0em, font=\scriptsize},
	x label style = {font=\small},
    y label style = {yshift=0.5em,font=\small},    
	axis y line*=right,
	enlargelimits=0.05,
    nodes near coords,
    every node near coord/.append style={black, font=\small, opacity=0.7, yshift=-0.em,xshift=0.0em },
	xticklabels={SPICE},
	xmax=7.5, xmin=0.5,
	% x tick label style = {xshift=-5em, yshift=0.05em, align=center,font=\small},
%	title=(a) BLEU-4,
	title style={yshift=-0.9em,font=\small},
	width = 8.5cm,
	height = 4cm,
	axis x line*=top,
	axis line style={-},
	xtick=\empty,
	]
	
	\addplot[c4, pattern= north east lines, thick, pattern color=c4, bar shift=0pt, bar width = 1.6em] coordinates
	{
		(5.34,  54.40)
	};
	\addplot[c3, pattern=crosshatch, pattern color=c3, thick, bar shift=0pt, bar width = 1.6em] coordinates
	{
		(6.66,  53.17)
	};
%	\addlegendentry{Pipeline}
%	\addplot[red, pattern=dots, pattern color=red, thick, bar shift=0pt, bar width = 1.6em] coordinates
%	{
%	(4.16,  32.58)
%	}
\end{axis}
\end{tikzpicture}
\vspace{-2mm}
\caption{Comparison between the results of scene graph features and object features.}
\label{fig:sg_obj}
\vspace{-3mm}
\end{figure}

% \pgfplotsset{compat=1.7,every axis title/.append style={at={(0.5,-0.7)}}}
\pgfplotsset{compat=1.7,every axis title/.append style={at={(0.5,-0.45)}, font=\fontsize{14}{1}\selectfont},every axis/.append style={xtick pos=left,ytick pos=left,tickwidth=1.5pt}}
\usetikzlibrary{matrix}
\usepgfplotslibrary{groupplots}
\usetikzlibrary{patterns,backgrounds}

\definecolor{c1}{RGB}{252,232,212}
\definecolor{c2}{RGB}{184,183,163}
\definecolor{c3}{RGB}{107,112,092}
\definecolor{c4}{RGB}{203,153,126}
\definecolor{c5}{RGB}{107,112,092}

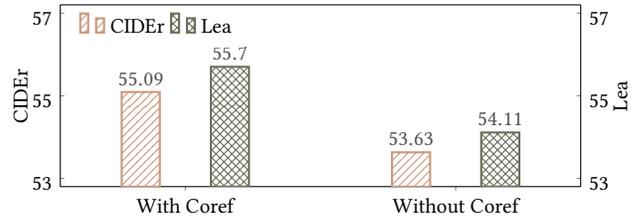
\begin{figure}[t]
\centering
\begin{tikzpicture}
%\begin{groupplot}[group style={group name=myplot,group size=2 by 1,horizontal sep=30pt,xlabels at=edge bottom, ylabels at=edge left},height=4cm,width=4cm]
\begin{axis}[
	ybar,
	ytick = {53, 55, 57},
	yticklabels={53, 55, 57},
	ymax=57,ymin=53,
	ylabel = CIDEr,
	y tick label style = {yshift=-0.5em, text height=0ex,font=\small},
	%    ytick style = {yshift=-0.0em, font=\scriptsize},
    x label style = {font=\small},
    y label style = {yshift=-0.5em, font=\small},
    axis x line*=bottom,
	axis line style={-},
    axis y line*=left,
	axis line style={-},
	enlargelimits=0.05,
    nodes near coords,
    every node near coord/.append style={black, font=\small, opacity=0.7, yshift=-0.1em,xshift=0.0em },
	legend style={at={(0.02,0.99)},anchor=north west, draw=none, legend columns=-1, font=\small},
	xticklabels={With Coref,Without Coref},
	xtick={2,6},
% 	ymajorgrids,
	xmax=7.5, xmin=0.5,
	x tick label style = {yshift=0.05em, align=center,font=\small},
%	title=(a) BLEU-4,
	title style={yshift=-0.9em,font=\small},
	width = 8.5cm,
	height = 4cm,
	]
	\addplot[c4, pattern= north east lines, thick, pattern color=c4, bar shift=0pt, bar width = 1.6em] coordinates
	{
		(1.34,  55.09)
	};\addlegendentry{CIDEr}
	
	\addplot[c3, pattern=crosshatch, pattern color=c3, thick, bar shift=0pt, bar width = 1.6em] coordinates
	{
		(2.66,  55.70)
	};\addlegendentry{Lea}
\end{axis}

\begin{axis}[
	ybar,
	ytick = {53, 55, 57},
	yticklabels={53, 55, 57},
	ymax=57,ymin=53,
	ylabel = Lea,
	y tick label style = {yshift=-0.5em, text height=0ex,font=\small},
%   ytick style = {yshift=-0.0em, font=\scriptsize},
	x label style = {font=\small},
    y label style = {yshift=0.5em,font=\small},    
	axis y line*=right,
	enlargelimits=0.05,
    nodes near coords,
    every node near coord/.append style={black, font=\small, opacity=0.7, yshift=-0.1em,xshift=0.0em },
	xmax=7.5, xmin=0.5,
	% x tick label style = {xshift=-5em, yshift=0.05em, align=center,font=\small},
%	title=(a) BLEU-4,
	title style={yshift=-0.9em,font=\small},
	width = 8.5cm,
	height = 4cm,
	axis x line*=top,
	axis line style={-},
	xtick=\empty,
	]
	
	\addplot[c4, pattern= north east lines, thick, pattern color=c4, bar shift=0pt, bar width = 1.6em] coordinates
	{
		(5.34,  53.63)
	};
	\addplot[c3, pattern=crosshatch, pattern color=c3, thick, bar shift=0pt, bar width = 1.6em] coordinates
	{
		(6.66,  54.11)
	};
%	\addlegendentry{Pipeline}
%	\addplot[red, pattern=dots, pattern color=red, thick, bar shift=0pt, bar width = 1.6em] coordinates
%	{
%	(4.16,  32.58)
%	}
\end{axis}
\end{tikzpicture}
\vspace{-2mm}
\caption{Comparison between the results of HostSG with and without cross-clip coreference edges.}
\label{fig:coref}
\vspace{-3mm}
\end{figure}

\begin{figure}[!t]
    \centering
    \includegraphics[width=0.98\linewidth]{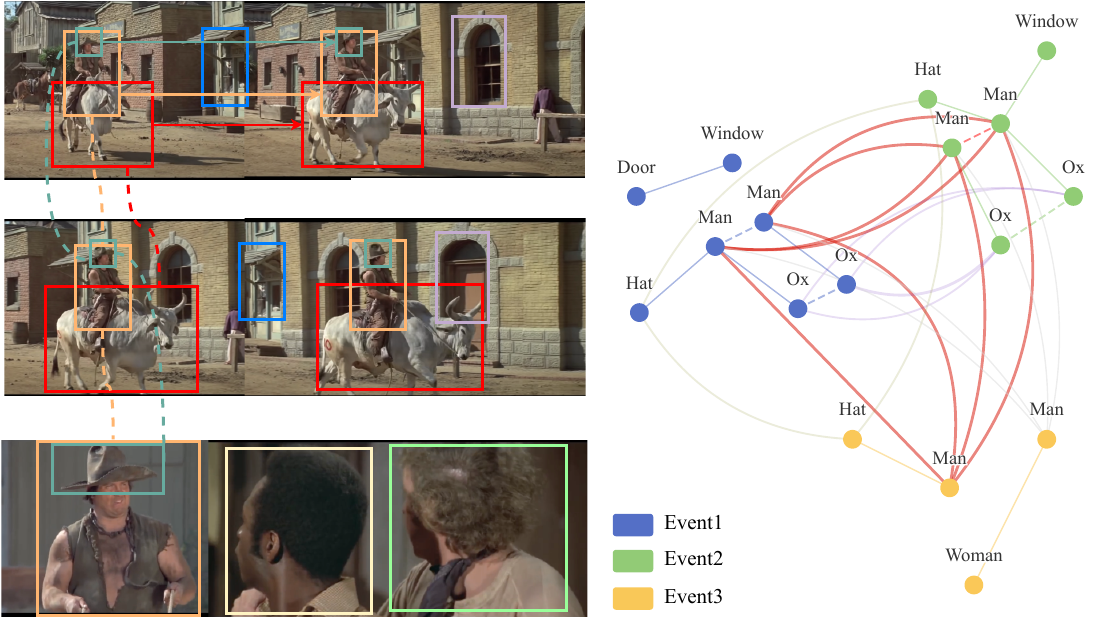}
    % \vspace{-3mm}
    \caption{
    Visualization of the cross-clip coreference edges. We select three clips from a video, representing the edge weights by the line width. The highlighted red lines denote the coreference relation of the objects with the tag ``Man''.
    }
    \label{fig:coref_case}
    % \vspace{-4mm}
\end{figure}

\paratitle{$\blacktriangleright$ \bf Q2: \textbf{How does the cross-clip coreference information aid the task?}}
Next, we consider investigating the influence of coreference information.
We cut all the cross-clip coreference edges of HostSG for comparison and keep the remaining structure, and then we can get the perception of its necessity.
The results are shown in Figure \ref{fig:coref}, the HostSG without cross-clip coreference decreases the performance by a noticeable margin.
We further provide a qualitative analysis to explore whether it correctly models the cross-clip coreference relations.
We consider the weights of cross-clip edges after the final layer. 
As shown in Figure \ref{fig:coref_case}, we can see the weights gather among the object nodes in a coreference cluster.

\begin{figure}[!t]
    \centering
    \includegraphics[width=0.98\linewidth]{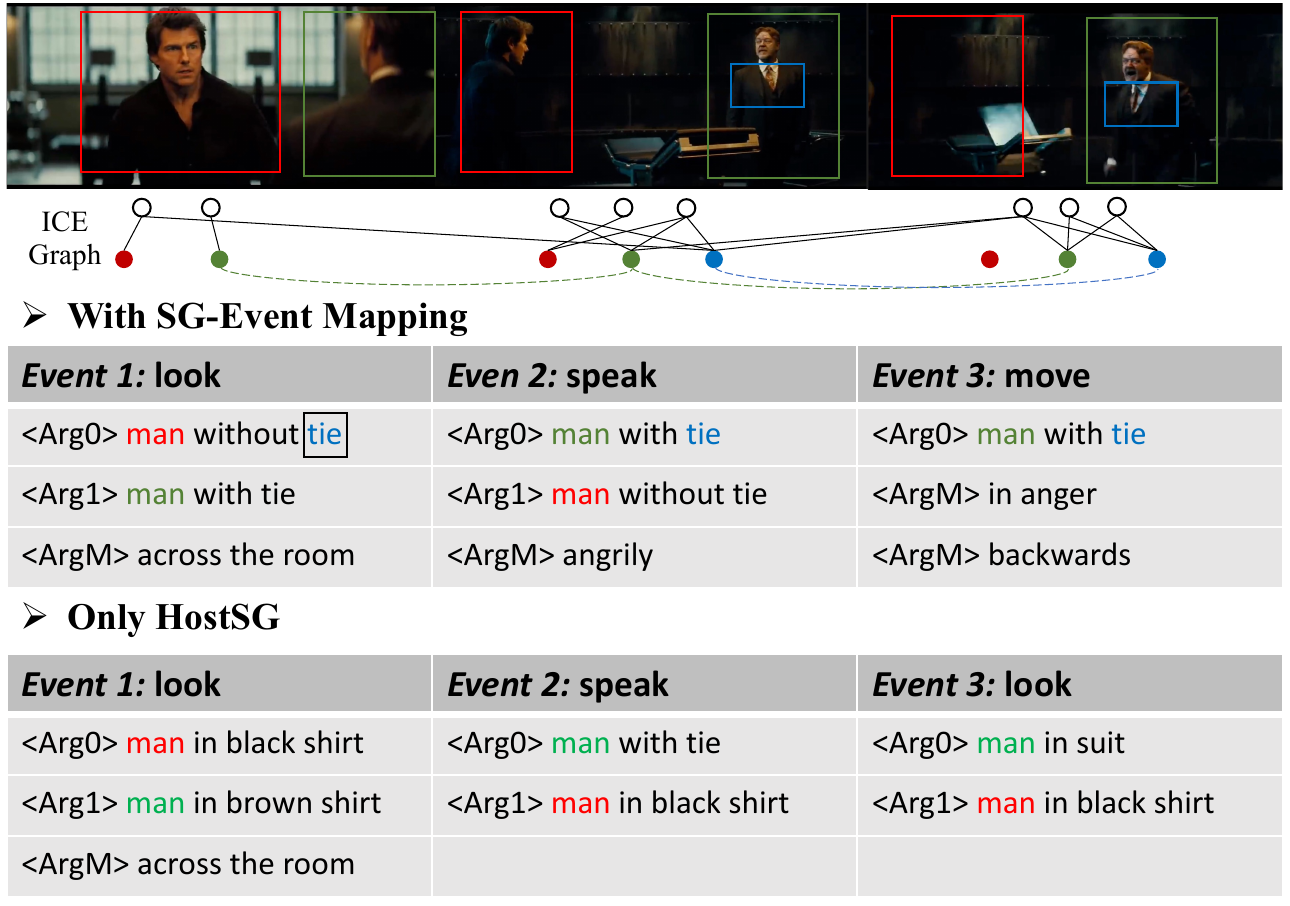}
    % \vspace{-3mm}
    \caption{
        Quantitative results of an example of our model with and without scene-event mapping.
    }
    \label{fig:case}
    % \vspace{-4mm}
\end{figure}

\paratitle{$\blacktriangleright$ \bf Q3: \textbf{Does the scene-event mapping mechanism really bridge the scene-event gap?}}
Overall, as shown in Table \ref{tab:ablation}, the scene-event mapping contributes significantly to the final performance, gaining 2.72 Acc@1 2.71 CIDEr and 1.16 Macro-Acc than the model with only HostSG features.
We can further empirically analyze the scene-event mapping via an example in Figure \ref{fig:case}.
In this case, the ICE graph structure reflects the scene-event mapping results.
In event 1, the model with only HostSG feature can not capture the cross-event object tie which is not presented in its own clip while the model with mapping does.
Furthermore, the model with only HostSG fails to predict the right predicate for event 3 by cross-event information (the state in event 2).
The case clearly demonstrates the effectiveness of the scene-event mapping mechanism. 
% \pgfplotsset{compat=1.7,every axis title/.append style={at={(0.5,-0.7)}}}
\pgfplotsset{compat=1.7,every axis title/.append style={at={(0.5,-0.45)}, font=\fontsize{14}{1}\selectfont},every axis/.append style={xtick pos=left,ytick pos=left,tickwidth=1.5pt}}
\usetikzlibrary{matrix}
\usepgfplotslibrary{groupplots}
\usetikzlibrary{patterns,backgrounds}

\definecolor{c1}{RGB}{252,232,212}
\definecolor{c2}{RGB}{184,183,163}
\definecolor{c3}{RGB}{107,112,092}
\definecolor{c4}{RGB}{203,153,126}
\definecolor{c5}{RGB}{107,112,092}

\begin{figure}[t]
\centering
\begin{tikzpicture}
\pgfplotsset{set layers=standard}
% \begin{groupplot}[group style={group name=myplot,group size=2 by 1,horizontal sep=30pt,xlabels at=edge bottom, ylabels at=edge left},height=9cm,width=8.5cm]
\begin{axis}[
	ybar,
	ytick = {10, 20, 30, 40, 50},
	yticklabels={10, 20, 30, 40, 50},
	ymax=53,ymin=10,
    xlabel = Training steps,
	ylabel = CIDEr,
	y tick label style = {yshift=-0.3em,xshift=0.1em, text height=0ex,font=\small,color=c4},
	%    ytick style = {yshift=-0.0em, font=\scriptsize},
    % x label style = {font=\small,xshift=-3.7cm,yshift=0.4cm,align=center},
    x label style = {font=\small},
    y label style = {yshift=-0.5em, font=\small,color=c4},
    axis x line*=bottom,
	axis line style={-},
    axis y line*=left,
	axis line style={-},
	enlargelimits=0.05,
    nodes near coords,
    every node near coord/.append style={black, font=\fontsize{4}{3.5}\selectfont, opacity=0.7, yshift=-0.2em,xshift=-0.3em },
	legend style={at={(0.02,0.99)},anchor=north west, draw=none, legend columns=-1, font=\small},
	xticklabels={0,500,1000,1500,2000,2500,3000,3500,4000,4500},
	xtick={1,2,3,4,5,6,7,8,9,10},
	% ymajorgrids,
	xmax=10.5, xmin=0.5,
	x tick label style = {yshift=0.05em, align=center,font=\small},
%	title=(a) BLEU-4,
	title style={yshift=-0.9em,font=\small},
	width = 8.25cm,
	height = 3.5cm,
    set layers,
	]
	\addplot[c1, fill=c1, bar shift=-0.3em, bar width = 0.6em] coordinates
	{
		(1,  11.02)
        (2,  17.48)
        (3,  26.94)
        (4,  33.29)
        (5,  38.64)
        (6,  41.30)
        (7,  44.93)
        (8,  46.20)
        (9,  48.01)
        (10, 48.92)
	};
\end{axis}

\begin{axis}[
	ybar,
	ytick = {10, 20, 30, 40, 50},
	yticklabels={10, 20, 30, 40, 50},
	ymax=53,ymin=10,
	ylabel = Lea,
	y tick label style = {yshift=-0.3em,xshift=-0.1em, text height=0ex,font=\small,color=c3},
	x label style = {font=\small},
    y label style = {yshift=0.5em,font=\small,color=c3},    
	axis y line*=right,
	enlargelimits=0.05,
    nodes near coords,
    every node near coord/.append style={black, font=\fontsize{4}{3.5}\selectfont, opacity=0.7, yshift=-0.1em,xshift=0.0em },
	xtick={1,2,3,4,5,6,7,8,9,10},
% 	ymajorgrids,
	xmax=10.5, xmin=0.5,
	% x tick label style = {xshift=-5em, yshift=0.05em, align=center,font=\small},
%	title=(a) BLEU-4,
	title style={yshift=-0.9em,font=\small},
	width = 8.25cm,
	height = 3.5cm,
	axis x line*=top,
	axis line style={-},
	xtick=\empty,
    set layers,
	]
	
	\addplot[c2, fill=c2, bar shift=0.25em, bar width = 0.5em] coordinates
	{
		(1,  20.70)
        (2,  29.32)
        (3,  35.33)
        (4,  42.27)
        (5,  45.01)
        (6,  47.37)
        (7,  48.64)
        (8,  49.12)
        (9,  49.83)
        (10, 50.95)
	};
%	\addlegendentry{Pipeline}
%	\addplot[red, pattern=dots, pattern color=red, thick, bar shift=0pt, bar width = 1.6em] coordinates
%	{
%	(4.16,  32.58)
%	}
\end{axis}

\begin{axis}
    [
	% yline,
    smooth,
	ytick = {20, 40, 60, 80, 100},
	ymax=100,ymin=-100,
	ylabel = {Edge Weights\\ Fluctuation (\%)},
    y label style = {yshift=-0.9em,xshift=4em,font=\small,align=center},
	y tick label style = {yshift=-0.5em,xshift=0.1em, text height=0ex,font=\small},
%   ytick style = {yshift=-0.0em, font=\scriptsize},
    xtick=\empty,
% 	ymajorgrids,
	xmax=10.5, xmin=0.5,
    ymajorgrids,
    grid style=dashed,
	% x tick label style = {xshift=-5em, yshift=0.05em, align=center,font=\small},
%	title=(a) BLEU-4,
	title style={yshift=-0.9em,font=\small},
    width = 8.25cm,
	height = 5.5cm,
    ]
	\addplot+[
        mark=*,
        mark options={c4,scale=0.5pt},
        c4,thick
    ] coordinates
	{
		(1,  70.37)
        (2,  50.28)
        (3,  47.63)
        (4,  41.18)
        (5,  37.09)
        (6,  33.26)
        (7,  20.71)
        (8,  19.80)
        (9,  16.38)
        (10, 15.99)
	};
%	\addlegendentry{Pipeline}
%	\addplot[red, pattern=dots, pattern color=red, thick, bar shift=0pt, bar width = 1.6em] coordinates
%	{
%	(4.16,  32.58)
%	}
\end{axis}
% \end{groupplot}

% \begin{axis}
%     [
% 	% yline,
%     smooth,
% 	ytick = {20, 40, 60, 80, 100},
% 	yticklabels={20, 40, 60, 80, 100},
% 	ymax=100,ymin=-100,
% 	ylabel = GIB Loss,
%     y label style = {yshift=1em,xshift=4.5em,font=\small},
% 	y tick label style = {yshift=-0.5em, text height=0ex,font=\small,color=c3},
% %   ytick style = {yshift=-0.0em, font=\scriptsize},
%     xtick=\empty,
% % 	ymajorgrids,
% 	xmax=10.5, xmin=0.5,
% 	% x tick label style = {xshift=-5em, yshift=0.05em, align=center,font=\small},
% %	title=(a) BLEU-4,
%     axis y line*=right,
% 	title style={yshift=-0.9em,font=\small},
%     width = 8.25cm,
% 	height = 7.5cm,
%     ]
% 	\addplot+[
%     ] coordinates
% 	{
% 		(1,  20.70)
%         (2,  29.32)
%         (3,  35.33)
%         (4,  42.27)
%         (5,  45.01)
%         (6,  47.37)
%         (7,  48.64)
%         (8,  49.12)
%         (9,  49.83)
%         (10, 50.95)
% 	};
% %	\addlegendentry{Pipeline}
% %	\addplot[red, pattern=dots, pattern color=red, thick, bar shift=0pt, bar width = 1.6em] coordinates
% %	{
% %	(4.16,  32.58)
% %	}
% \end{axis}

\end{tikzpicture}
\vspace{-7mm}
\caption{The trends of changing amplitude of edge weights (average of all adjusted edges), along with task performance.}
\label{fig:refinement}
\vspace{-1mm}
\end{figure}
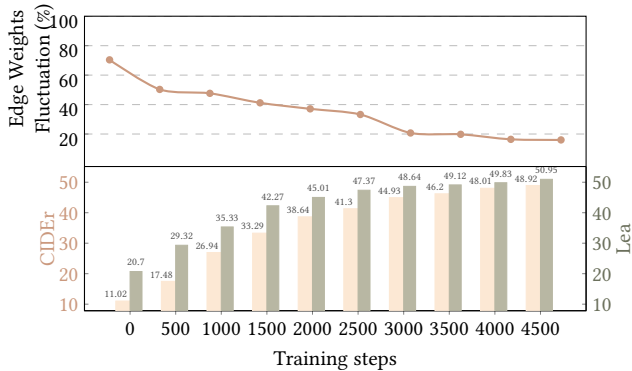

\paratitle{$\blacktriangleright$ \bf Q4: \textbf{Can iterative refinement really denoise the edges?}}
First, we plot the trajectories of edge-adjusting amplitudes through training steps, along with the trends of task performance.
As shown in Figure \ref{fig:refinement}, the weight variability of edges decreases gradually while the task performance climbs along with the training process.
The chart shows that the edge refinement indeed optimizes the graph structure by adjusting edge weights.

\paratitle{{$\blacktriangleright$ \bf Q5: \textbf{Impact of the temporalized dynamic scene graph}.}}
In realistic scenarios, we expect to model each object in each frame for video understanding.
But in practice, there are too many redundant objects in frames with a high sampling rate, which will cause extreme computational costs.
The proposed temporalized dynamic scene graph could compress the DSG by merging the static nodes.
We could assume that their slight movement in a short clip will not change their semantics.
This is reasonable because most static objects in a video scene provide little information to understand the salient event. 
Table \ref{tab:tsg} gives the comparison between a model with TSG and raw DSG, where the performance in the two situations is consistent.
That means the static node merging does not lose key information.
We can also find how the object number affects the performance.
The results show that the fewer objects may lose some information while the redundant objects may bring noisy information.
\begin{table}[t]
    \fontsize{9}{11.5}\selectfont
    \setlength{\tabcolsep}{1.8mm}
    \begin{center}
    % \resizebox{1\textwidth}{!}{
    \caption{Comparison between TSG and raw DSG without node merging, which means we treat all the DSG nodes as dynamic nodes. ``Obj Num'' means keep top N objects when parsing the scene graph.}
    \begin{tabular}{lcccc}
    \hline
    \multicolumn{1}{c}{} & \bf Obj Num & \bf Acc@1 & \bf CIDEr & \bf Macro-Acc \\
    % \hline
    % \multicolumn{4}{l}{\textbf{2D vs. 3D Info}} \\
    % & 2D & - & - \\
    % & 3D & - & - \\
    \hline
    \multirow{3}{*}{Raw DSG} & 5 & 54.19 & 53.03 & 34.91 \\
    & 15 & 56.21 & \bf 55.10 & 35.88 \\
    & 20 & 55.53 & 54.91 & 34.80 \\
    \hline
    \multirow{3}{*}{TSG} & 5 & 54.31 & 52.77 & 34.66 \\
    & 15 & \bf 56.15 & 55.09 & \bf 35.97 \\
    & 20 & 55.38 & 54.20 & 35.47 \\

    \hline
    \end{tabular}
    % }
    \vspace{-2mm}
    \label{tab:tsg}
    \end{center}
    \vspace{-1mm}
    \end{table}

    % For ``w/o \textsc{Go3D}-S$^2$G'', we 
    
    % removing the graph feature $r^{G}$ and feed the embedded 3D features of target objects $\bm{s}_i^v$ in Equation \ref{equation:v} to the encoder for text generation.
    % For ``w/o O\textsc{C}GCN'', we replace the O\textsc{C}GCN with a normal GCN.
    % For ``w/o S$^3$ mechanism'', we using the average-pooling results of all the nodes in the scene graph as graph feature for text generation.
\section{Conclusions}
In this paper, we introduce a novel approach to the challenging VidSRL task.
Our approach involves modeling the fine-grained spatial semantics and temporal dynamics of videos by constructing a holistic spatio-temporal scene graph representation. 
We devise a scene-event mapping mechanism to bridge the gap between the underlying scene structure and the high-level event semantic structure.
Furthermore, we propose an iterative structure refinement strategy 
% that can iteratively update the edge weights 
to enhance the compatibility between the graph representations and VidSRL reasoning.
% , ensuring that the resulting graph representations are best suited for VidSRL reasoning.
The proposed framework enables us to jointly predict the three subtasks of VidSRL, which effectively avoids the error propagation issue in the current pipeline paradigm.
Empirical evaluations on the benchmark dataset demonstrate that our approach outperforms current methods.

\bibliography{custom}
\bibliographystyle{ACM-Reference-Format}

\clearpage
\appendix

\section{Detailed GIB-guided Assistant Optimization}
Here, we provide more background information about the graph information bottleneck (GIB) principle.
Given the original graph G, and the target Y, the goal of representation learning is to obtain the refined graph $G^-$ which is maximally informative w.r.t Y:
\begin{equation}\label{eq:19}
    \min_{G^-}[-I(G^-,Y)+\beta I(G^-,G)],
\end{equation}
where $I(G^-, G)$ minimizes the mutual information between $G$ and $G^-$ such that $G^-$ learns to be the minimal and compact one of $G$.
$I(G^-, Y)$ is the prediction objective, which encourages $G^-$ to be informative enough to predict the label $Y$.
$\beta$ is a Lagrangian coefficient.
For our HostSG, we denote $\bm{z}$ as the compact information of the resulting $G^-$, which is sampled from a Gaussian distribution parameterized by $\bm{a}_v$ and $\bm{a}_{rel}$, which are obtained by Eq. (14).
The the Eq. \ref{eq:19} can be rephrased as:
\begin{equation}
    \mathcal{L}_{GIB} = \min[-I(\bm{z}_v,Y_v)-I(\bm{z}_{rel},Y_{rel})+I(\bm{z}_v, G)+I(\bm{z}_{rel}, G)],
\end{equation}
where $Y_v$ and $Y_{rel}$ are the labels of verb prediction and event relation prediction.
The term $-I(\bm{z}_v, Y_v)$ and $-I(\bm{z}_{rel},Y_{rel})$ can be expanded as:
\begin{equation}
\begin{split}
    -I(\bm{z}_v,Y_v)&\leq -\int p(Y_v,\bm{z}_v)\log q(Y_v|\bm{z}_v)dY_v d\bm{z}_v+H(Y_v)\\
    &:=\mathcal{L}_{CE}(q(Y_v|\bm{z}_v),Y_v),\\
     -I(\bm{z}_{rel},Y_{rel})&\leq -\int p(Y_{rel},\bm{z}_{rel})\log q(Y_{rel}|\bm{z}_{rel})dYd\bm{z}+H(Y_{rel})\\
    &:=\mathcal{L}_{CE}(q(Y_{rel}|\bm{z}_{rel}),Y_{rel}),
\end{split}
\end{equation}
where $q(Y_v|\bm{z}_v)$ and $q(Y_{rel}|\bm{z}_{rel})$ are variational approximation of the true posterior $p(Y_v|\bm{z}_v)$ and $p(Y_{rel}|\bm{z}_{rel})$.
For $I(\bm{z}_v,G)$ and $I(\bm{z}_{rel},G)$, we estimate their upper bound via reparameterization trick:
\begin{equation}
\begin{split}
    I(\bm{z}_v,G)&\leq\int p(\bm{z}_v|G)\log\frac{p(\bm{z}_v|G)}{r(\bm{z}_v)}d\bm{z}dG\\
    &:=\text{KL}(p(\bm{z}_v|G)||r(\bm{z}_v)),\\
    I(\bm{z}_{rel},G)&\leq\int p(\bm{z}_{rel}|G)\log\frac{p(\bm{z}_{rel}|G)}{r(\bm{z})}d\bm{z}dG\\
    &:=\text{KL}(p(\bm{z}_{rel}|G)||r(\bm{z}_{rel})),
\end{split}
\end{equation}

\section{Detailed Implementation Settings}
% \subsection{Dataset}
% \input{tables/dataset}
% We evaluate our model on the benchmark dataset VidSitu, providing extensive verb and argument structure annotations for more than 130k video clips. Each video clip is approximately two seconds in length with 10 verbs annotated in the evaluation split, and one verb annotated in the training split. There are verb-senses defined in the event ontology, and each verb has at least 3 semantic roles. Table \ref{tab:data} shows the dataset statistics. Table \ref{tab:args} lists the pre-defined arguments. 
% \input{tables/arguments}

\begin{figure}[t]
    \centering
    \includegraphics[width=0.98\linewidth]{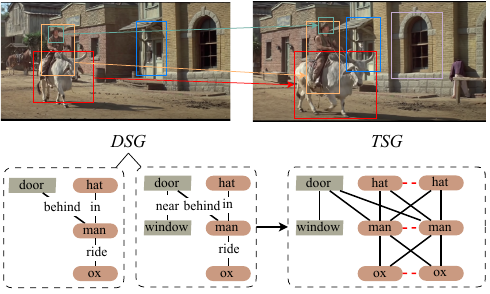}
    % \vspace{-2mm}
    \caption{
    Process of DSG merging.}
    \label{fig:app_tsg}
    % \vspace{-4mm}
\end{figure}
% \subsection{Model Hyperparameters}
\begin{table}[t]
    \fontsize{8}{11}\selectfont
    \setlength{\tabcolsep}{1.mm}
    \begin{center}
    % \resizebox{1\textwidth}{!}{
    \caption{Model hyperparameters.}
    \vspace{-3mm}
    \begin{tabular}{lc}
    \hline
     \bf Hyper-param. & \bf Value \\
    \hline
    dimension of ROI feature & 2048 \\
    dimension of Transformer hiddens & 1024 \\
    dimention of GAT hiddens & 1024 \\
    dimention of GGNN hiddens & 1024 \\
    layer number of GAT & 3 \\
    refinement times each batch & 2 \\
    max object number & 15 \\
    IoU threshold $d$ & 0.5 \\
    similarity threshold $\gamma$ & 0.3 \\
    S$3$ cutting threshold $p_{cut}$ & 0.1 \\
    max text length & 80 \\
    optimizer & Adam \\
    dropout & 0.1 \\
    learning rate & 1e-4 \\
    bach size & 16 \\
    epoch & 20 \\
    beam search number & 5 \\

    \hline
    \end{tabular}
    % }
    \label{tab:hyperparameter}
    \end{center}
    \end{table}
A summary of the hyperparameter choice is shown in Table \ref{tab:hyperparameter}

\section{Model Details}
% \subsection{Scene Graph Generation}
% We follow the prior practice \cite{DBLP:journals/corr/abs-2304-00733} to generate the scene graph.
% We employ the MaskRCNN as the object detector to obtain all the object nodes, and use MOTIFS as a relation classifier to obtain the relation labels (nodes) as well as the relational edges, which is pre-trained using the Visual Genome (VG) dataset.
% The dataset has 150 object labels and 50 relation labels.
% % Figure \ref{fig:tag_dist} shows the top 20 objects in the VidStu dataset.
% We filter the nodes and edges by sorting theme by scores and keeping top N ones.

\begin{figure}[t]
    \centering
    \includegraphics[width=0.98\linewidth]{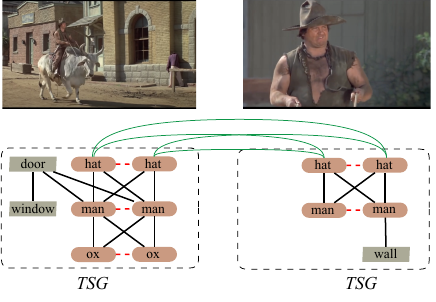}
    % \vspace{-2mm}
    \caption{
    Augmented holistic event-arguments semantic graph.}
    \label{fig:app_host}
    % \vspace{-4mm}
\end{figure}
\begin{table}[t]
    \fontsize{8}{11}\selectfont
    \setlength{\tabcolsep}{1.mm}
    \begin{center}
    % \resizebox{1\textwidth}{!}{
    \caption{Static and dynamic objects.}
    \vspace{-3mm}
    \begin{tabularx}{\linewidth}{lX}
    \hline
     \bf Type & \bf Object Labels \\
    \hline
    static object & bed,bag, banana, basket, beach, bench, board, bottle, bowl, box, branch, building, cabinet, chair, clock, counter, cup, curtain, desk, door, drawer, fence, flag, flower, food, fork, fruit, glass, handle, hill, house, lamp, laptop, leaf, letter, light, logo, mountain, number, orange, paper, phone, pillow, pizza, plant, plate, pole, post, pot, racket, railing, rock, roof, room, screen, seat, shelf, sidewalk, sign, sink, skateboard, ski, snow, stand, street, surfboard, table, tile, tire, toilet, towel, tower, track, tree, trunk, umbrella, vase, vegetable, window, windshield, wire\\
    \hline
    dynamic object & airplane, arm, animal, bear, bird, bike, boat, boot, boy, bus, cap, car, cat, child, coat, cow, dog, ear, elephant, engine, eye, face, finger, girl, giraffe, glove, guy, hair, hand, hat, head, helmet, horse, jacket, jean, kid, kite, lady, leg, man, men, motorcycle, mouth, neck, nose, pant, paw, people, person, plane, player, sheep, shirt, shoe, short, skier, sneaker, sock, tail, tie, train, truck, vehicle, wave, wheel, wing, woman, zebra \\

    \hline
    \end{tabularx}
    \label{tab:obj}
    \end{center}
    \end{table}
\subsection{Detailed process of HostSG Construction}
% Here we introduce the detailed process of HostSG construction.
% \paragraph{\bf Process of DSG Merging.}
We divide the DSG nodes into static and dynamic ones and merge the dynamic ones for compressing.
Though the visual appearance of a static may change from frame to frame, we assume that their semantic change due to the small interval between keyframes (about 0.4s in our implementation).
However for dynamic nodes, such as a {\it person}, we could not ignore their semantic changing due to their frequent interaction with various objects.
Thus we keep the dynamic nodes and add motions edges to capture these interaction, merging other static nodes for compressing.
Concretely, we process the DSG of each frame one by one from the beginning of a clip.
Figure \ref{fig:app_tsg} shows an example of the DSG merging of two frames, where the ``door'' and ``window'' are the static nodes, the ``man'', ``hat'' and ``ox'' are the dynamic nodes.
We merge the ``door'' nodes that exists in both frames and add motion edges between nodes of ``hat'', ``man'' and ``ox'' in the two frame separately, keeping their connection with other nodes.
The merging is constrained by the object label and IoU (Eq. 2).
Then we continue this process with subsequent frames of one clip to obtain the TSG graph finally.
The static and dynamic objects are shown in Table \ref{tab:obj}

For each clip, we construct a TSG graph, and with the TSGs in hand (5 TSGs each sample in VidSRL data), we combine these TSGs into the HostSG by adding cross-clip co-reference edges.
The co-reference edge construction is constrained by object label and similarity score (Eq. 3,4).
When processing dynamic nodes, we should add full-connection edges in a co-reference cluster, as shown in Figure \ref{fig:app_host}.

\begin{figure}[t]
    \centering
    \includegraphics[width=0.98\linewidth]{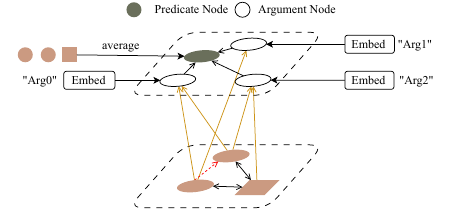}
    % \vspace{-2mm}
    \caption{
    Event nodes generation.}
    \label{fig:app_ice}
    % \vspace{-4mm}
\end{figure}

\begin{figure}[t]
    \centering
    \includegraphics[width=0.98\linewidth]{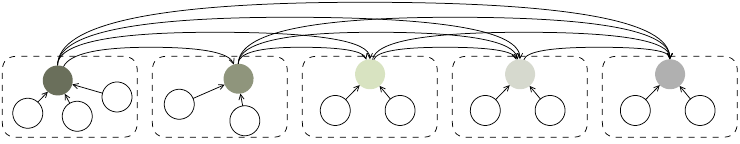}
    % \vspace{-2mm}
    \caption{
    Edges between predicate nodes.}
    \label{fig:app_pre_node}
    % \vspace{-4mm}
\end{figure}
\subsection{Event Nodes Generation in scene-event mapping}
Here we introduce the details of event nodes generation.
As shown in Figure \ref{fig:app_ice}, we generate a predicate node and some argument nodes for each TSG.
In our implementation, we generate 5 real argument nodes and 5 modifier argument nodes for each TSG.
The initial representation of the predicate node is the average of all the TSG nodes in a clip, and the initial representation of the argument nodes is the embedding of the argument tags.

The we create edges between predicate nodes across all events, in which each predicate node directionally points to all the subsequent ones.
As show in Figure \ref{fig:app_pre_node}, the structure of predicate nodes is a half full-connection, indicating the sequential occurring of events.

\begin{figure}[t]
    \centering
    \includegraphics[width=0.98\linewidth]{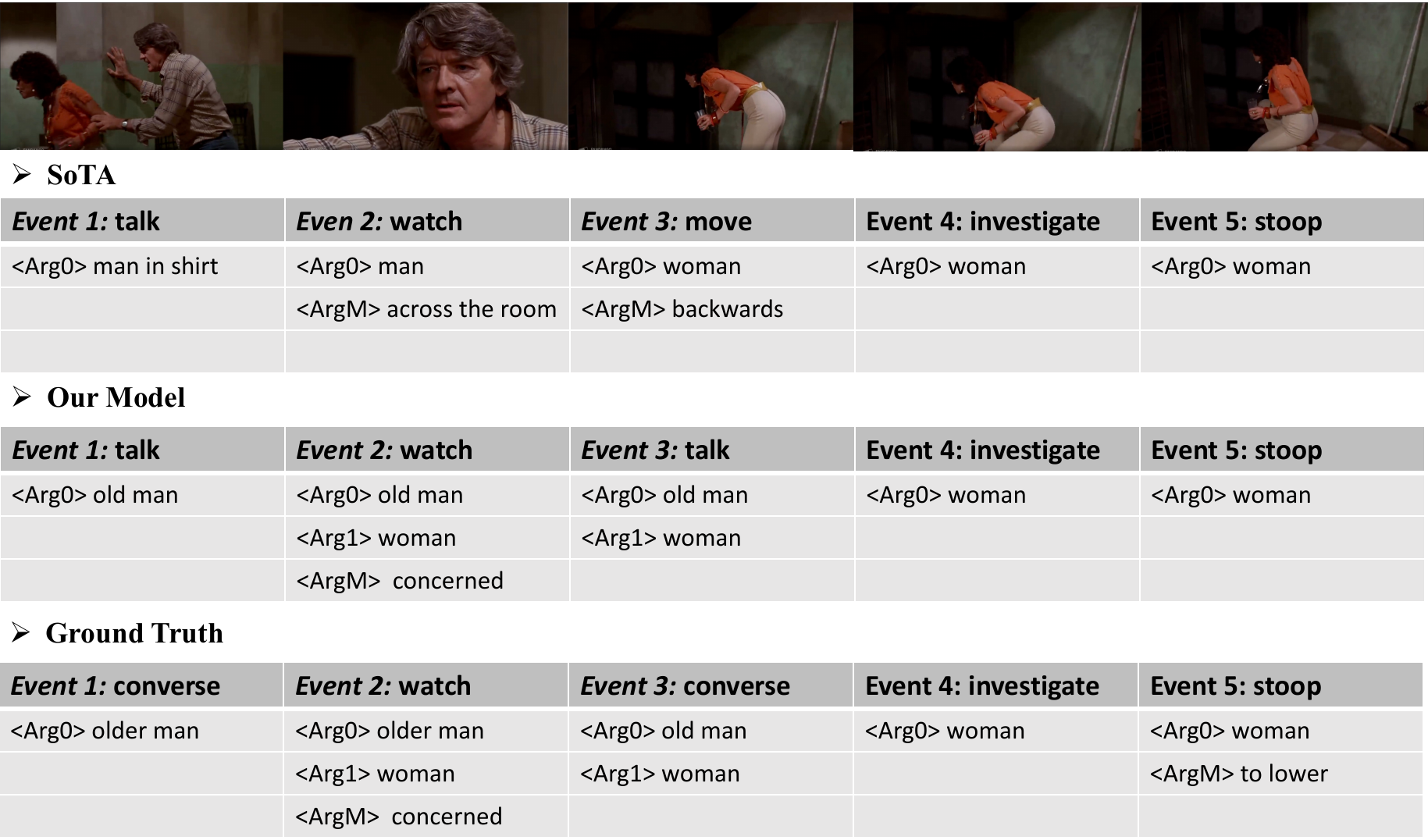}
    \includegraphics[width=0.98\linewidth]{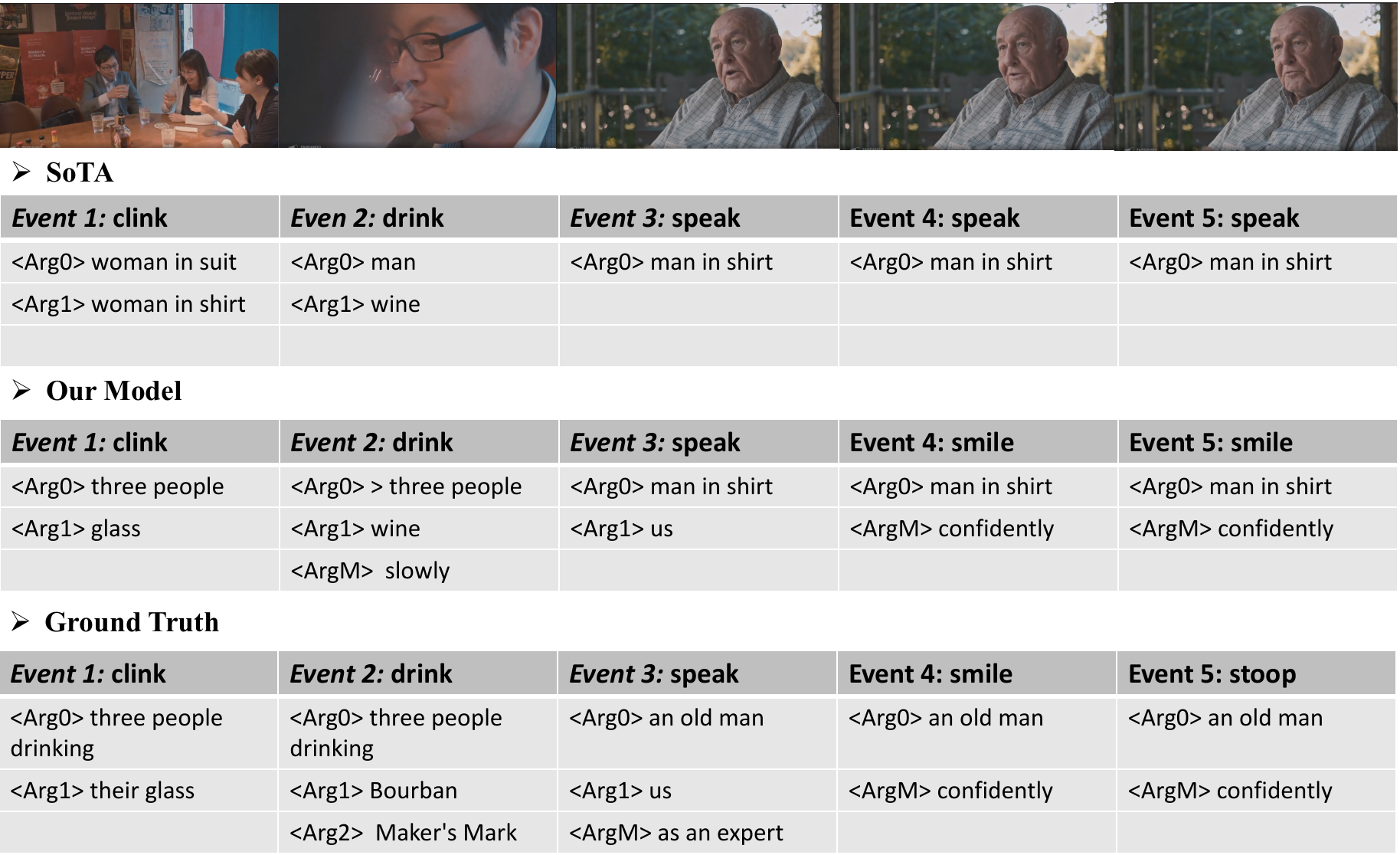}
    % \vspace{-2mm}
    \caption{
    Results comparison among current SoTA method, our model and the ground truth of verbs and roles.}
    \label{fig:case_app}
    % \vspace{-4mm}
\end{figure}
\section{Case Study}
We show more case studies here.
As shown in figure \ref{fig:case_app}, our model could capture the cross-event information while the current SoTA model cannot.
% If the first case, in event 2, the ``man'' is watching the woman who is not seen in this clip, and our model cold recognize this relation.
% In the second case, we can see our model predict more accurate verbs and generate more informative arguments through modeling information from holistic scene.

\end{document}